\documentclass{article} 
\usepackage{iclr2027_conference,times}

\usepackage{amsmath,amsfonts,bm}

\def\eqref#1{equation~\ref{#1}}

\def\1{\bm{1}}

\DeclareMathAlphabet{\mathsfit}{\encodingdefault}{\sfdefault}{m}{sl}
\SetMathAlphabet{\mathsfit}{bold}{\encodingdefault}{\sfdefault}{bx}{n}

\usepackage{hyperref}
\usepackage{url}

\usepackage{ours}

\title{Text-to-Image Models Need Less from Text Encoders Than You Think}

\author{%
  Nurit Spingarn$^{*1}$\quad
  Noa Cohen$^{*1}$ \quad 
  Tamar Rott Shaham$^{2}$ \quad 
  Tomer Michaeli$^{1}$ \\[1pt]
  $^{1}$ Technion -- Israel Institute of Technology \quad
  $^{2}$ MIT CSAIL \\
}

\iclrfinalcopy 
\begin{document}


\maketitle

\begin{abstract}
Text-to-image models rely on text prompts as their primary interface to human intent. Prompts are encoded by a text encoder into embeddings that condition the image generation process. Beyond individual token meanings, text embeddings encode contextual information across the full prompt, such as compositionality and attribute binding. However, whether image models actually exploit this richer information remains underexplored. Here, we address the question: Which aspects of text representation are essential for image generation? We show that text-to-image diffusion transformer-based models commonly rely only on two relatively straightforward aspects of text representations: (i)~the merging of adjacent tokens into a word representation, for words spanning multiple tokens, and (ii)~word order, which is imprinted by the positional embedding of the text-encoder. To show this, we construct a new text embedding that encodes only individual word meanings and order but lacks any contextual information about the full prompt. We find that this \emph{bag of position-tagged words} representation is sufficient to successfully guide image generation, achieving visual quality and text fidelity that are on par with full text embedding-guided generation. This demonstrates that, contrary to common belief, text-to-image models often do not use the rich information encoded in the text embedding beyond individual word meanings and word order. Instead, the decoding of complex linguistic structures is performed by the image model itself.

\end{abstract}

\section{Introduction}

Text-to-image (TTI) models have seen remarkable progress in recent years, with systems now capable of producing highly realistic and semantically rich images from complex natural language prompts. Central to these systems is the text encoder, which serves as the interface between the user's intent, which is described by a text prompt, and the text embedding that conditions the image model. As generation capabilities advanced, models adopted increasingly powerful text encoders. Early systems relied on CLIP-based conditioning~\citep{radford2021clip}, while later models, such as Imagen~\citep{saharia2022photorealistic}, introduced large pretrained language models, like T5~\citep{raffel2020t5}, significantly improving generation quality on benchmarks such as DrawBench~\citep{saharia2022photorealistic}. More recent systems extend this trend: models such as Stable Diffusion (SD)~3~\citep{esser2024sd3} combine multiple text encoders, while architectures like FLUX.2~\citep{flux2_klein} incorporate large language models such as Qwen~\citep{yang2025qwen3} as encoders. This progression reflects a widely held assumption: richer and more expressive text representations lead to improved image generation.

\begin{figure}[ht]
    \centering
    \includegraphics[width=\textwidth]{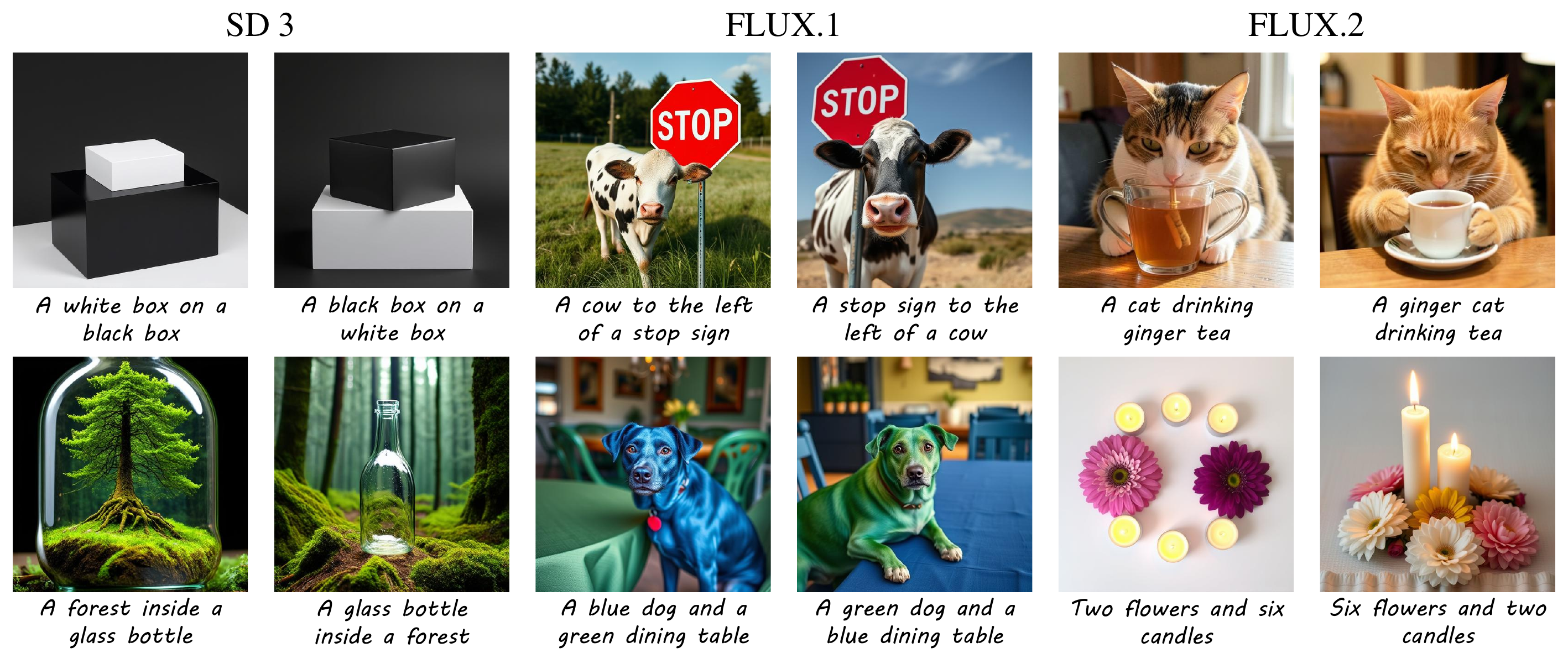}
    \caption{
    \textbf{Contextless text embeddings are often enough.} We find that when pretrained TTI models are conditioned on text embeddings that are stripped of any contextual information, they maintain high visual quality and prompt adherence. This surprising behavior is exhibited even for complex prompts that involve attribute binding, spatial relations, and numeracy. We show that the capability of generating text-adherent images from a bag-of-words-like textual representation is enabled by the positional information encoded within each word embedding. This information allows the TTI model to indirectly decipher the word's role within the prompt and to disambiguate between different sentences that comprise the same set of words, as illustrated above. Our observations suggest that TTI models extract less information from the text embeddings than is commonly thought.
    }
    \label{fig:teaser}
\end{figure}

Despite this progress, it remains unclear which aspects of these rich text representations are actually utilized by the image generation models. Modern text encoders are capable of capturing complex linguistic structure, including compositionality, attribute binding, and long-range dependencies across the input prompt~\citep{tenney2019bert, jawahar2019does, hewitt2019structural}.
However, the presence of such information in the text representation does not necessarily imply that it is functionally used by the image model. Here, we ask: which aspects of text representations are essential for guiding image generation?

\begin{wrapfigure}{r}{0.5\textwidth}
    \centering
    \captionsetup[subfigure]{labelformat=simple}
    \renewcommand{\thesubfigure}{($\roman{subfigure}$)}
    \begin{subfigure}{0.16\textwidth}
        \centering
        \includegraphics[width=\linewidth, valign=t]{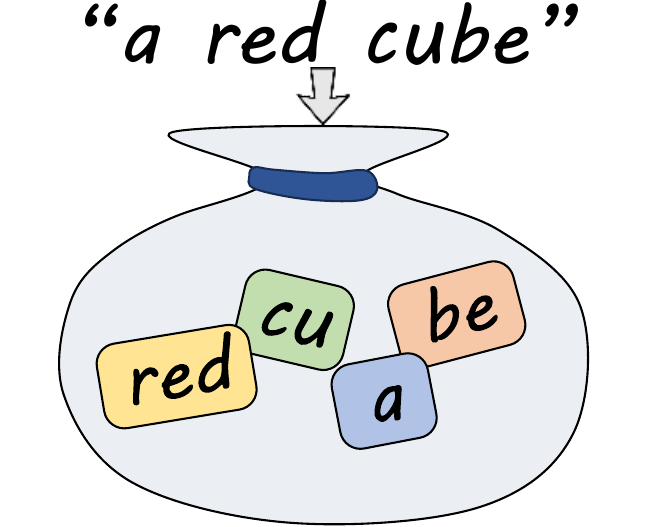} 
        \caption{BoT}
        \label{fig:bot_bag}
    \end{subfigure}
    \begin{subfigure}{0.16\textwidth}
        \centering
        \includegraphics[width=\linewidth, valign=t]{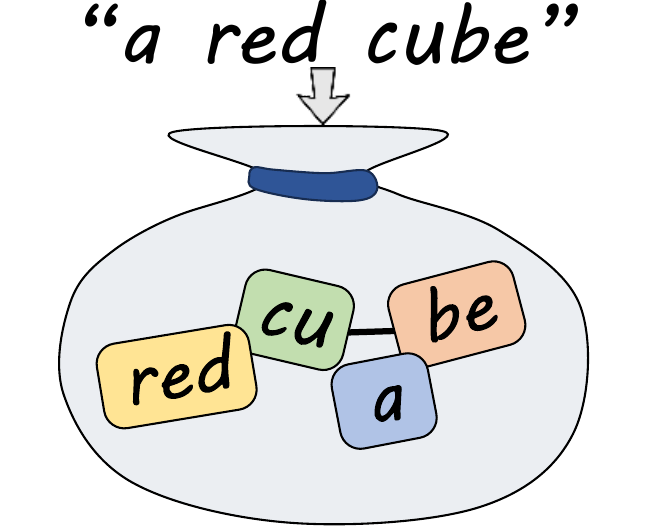}
        \caption{BoW}
        \label{fig:bow_bag}
    \end{subfigure}
    \begin{subfigure}{0.16\textwidth}
        \centering
        \includegraphics[width=\linewidth, valign=t]{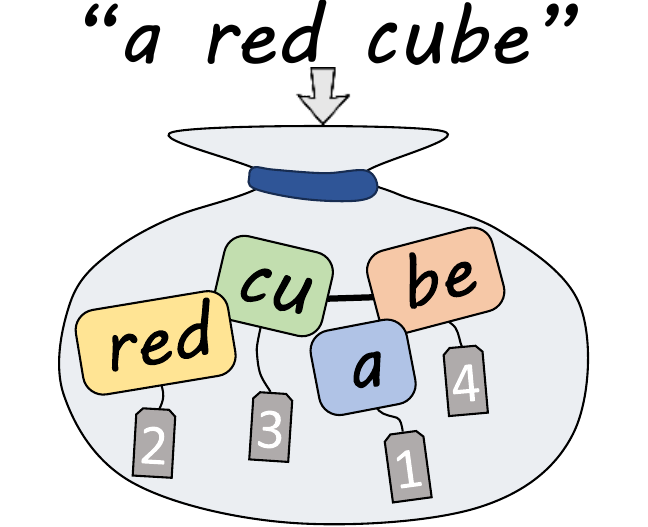}
        \caption{BoPTW}
        \label{fig:boptw_bag}
    \end{subfigure}
    \caption{\textbf{Illustration of our three contextless text embeddings.}
    We propose three embedding types of increasing richness: ($i$) Bag of Tokens, where each token is represented independently; ($ii$) Bag of Words, where tokens are merged into word-level representations; and ($iii$) Bag of Position-Tagged Words, where word embeddings additionally reflect their position in the prompt.
    }
    \label{fig:bags}
\end{wrapfigure}

To answer this, we construct modified text embeddings that retain only specific aspects of the text prompt representation but contain no contextual information. We focus on three contextless text embedding types, as illustrated conceptually in~\cref{fig:bags}: ($i$)~Bag-of-Tokens (BoT), where the embedding contains information about each token separately, without additional context from the full prompt, ($ii$)~Bag-of-Words (BoW), where we represent words (that might be composed of multiple tokens) similarly without full-prompt context, and ($iii$)~Bag-of-Position-Tagged-Words (BoPTW), where we let the word embeddings in the bag of words vary according to their order in the prompt. 
Importantly, our contextless embeddings serve as drop-in replacements for the text encoder output, requiring no modifications to the pretrained image model. We construct them by replacing each token's full embedding with an average over its embeddings in unrelated sentences, carefully controlling included sentences to isolate specific aspect of the representation.

We study how pretrained transformer-based diffusion models react to our contextless embeddings, focusing on SD~3~\citep{esser2024sd3}, FLUX.1~Schnell~\citep{flux2024}, and FLUX.2~Klein-4B~\citep{flux2_klein}. We use prompts from the DrawBench~\citep{saharia2022photorealistic}, GenEval~\citep{ghosh2023geneval}, and the MS-COCO~2014 validation set~\citep{lin2014microsoft}, which cover diverse image generation aspects like compositionality, counting, color fidelity, and attribute binding. 
The generated images are assessed using a VLM as a judge, a well established methodology for evaluating image generation~\citep{zheng2023judging,do2026vision,peng2024dreambench++,lee2024prometheus,lee2024vhelm}. 
Comparing image generation conditioned on our simplified representations to generation with full text embeddings, reveals which aspects of text representations are required to guide image synthesis.

We find that all contextless embedding variants achieve surprisingly strong results. For simple prompts, a large fraction of the cases require only BoT representations to achieve good results, suggesting that individual token meanings alone are often sufficient to guide generation. For complex prompts, the best results are achieved with the BoPTW embeddings. 
Despite lacking compositional structure, attribute binding, and other forms of contextual interaction between words, BoPTW achieves results that are on par with the full text embedding across a range of challenging scenarios.
These include object counting, spatial relationships and attribute binding, as seen in Fig.~\ref{fig:teaser}.

These results demonstrate that, contrary to common belief, TTI models often do not use the rich contextual information encoded in text embeddings beyond individual word meanings and their order.
This calls for a reframing of our understanding of the respective roles of the text encoder and image model, and raises questions about the prevailing drive toward increasingly large and complex text encoders. It suggests that future text-to-image architectures may benefit from focusing on the image model's capacity to interpret linguistic structure, and on developing more efficient text representations tailored to what image models actually need.
\vspace{-4pt}
\section{Related Work}
\vspace{-4pt}
\myparagraph{Text conditioning in TTI models.}
Contemporary text-to-image generation is dominated by latent diffusion and flow-based architectures that condition image synthesis on pretrained language representations. 
Imagen~\citep{saharia2022photorealistic} was among the first models to demonstrate the critical role of large-scale text encoders, showing that T5~\citep{raffel2020t5} ($\sim11$B parameters) and CLIP~\citep{radford2021clip} ($\sim124$M parameters) text embeddings substantially improve compositional reasoning and performance on complex prompts.
Subsequent diffusion-based models, including SD~\citep{rombach2022high, esser2024sd3, podell2023sdxl} and FLUX.1~\citep{flux2024} adopted a combination of the T5 and CLIP  text encoders for further improving the representation of the text embeddings and improve the image generation capabilities.
More recent models, like FLUX.2~\citep{flux2_klein}, replace traditional encoders with large language models such as Mistral \citep{jiang2023mistral} ($\sim 24$B parameters) and Qwen~\citep{yang2025qwen3} ($\sim 30$B  parameters).
These approaches share a common architectural paradigm: the text encoder is pretrained independently of the image generator, and its frozen representations serve as the primary interface between language and visual synthesis. This design implicitly assumes that the richness of these representations (\eg their contextual, compositional, and relational structure) is effectively used by the image model.
\citet{wang2025scaling} suggest that text encoders in such models are often overparameterized. However, the precise requirements of the image model from the text embedding remain underexplored. Here, we systematically study this question by stripping off contextual information from the embeddings and analyzing the resulting image generation behavior. 

\myparagraph{Interpreting the role of text conditioning in TTI models.}
While much work has focused on interpreting cross-attention mechanisms in diffusion models \citep{hertz2022prompt,tang2023daam,chefer2023attend,ventura2026deleaker}, fewer studies have explicitly examined the \textit{text encoder} itself within TTI pipelines. 
\citet{toker2024diffusionlens} analyze intermediate states of the text encoder by observing the images generated when feeding them to the diffusion model.
Our approach complements this by showing that, in many practical settings, contextual structure can be discarded entirely without significantly degrading generation quality. 
Complementary to this, \citet{park2024emergence} investigate learning dynamics in concept space and 
show that high-level semantic capabilities can emerge during training in a non-linear fashion, suggesting that compositional structure emerges throughout the generative process.
\citet{wang2026circuit} provide a mechanistic, circuit-level analysis of diffusion models, identifying structured internal pathways responsible for spatial and relational reasoning. 
This supports the view that compositional computation is implemented within the generative model itself, rather than being fully expressed in the text embedding.

\myparagraph{Linguistic structure in CLIP.}
Despite their widespread adoption, CLIP-style encoders encode limited relational and compositional structure. The original CLIP work~\citep{radford2021clip} already noted bag of words like behavior, with sensitivity to individual concepts but not to their relationships. Subsequent works showed that CLIP is largely insensitive to word order~\citep{yuksekgonul2022and}, struggles with compositional distinctions~\citep{thrush2022winoground}, and exhibits invariance to relations and ordering~\citep{yuksekgonul2023aro}, even under hard negatives~\citep{hsieh2023sugarcrepe}. 
Here, we show that despite transitioning to text encoders that are more advanced than CLIP, TTI models often do not exploit the rich information these encoders provide. This suggesting that much of the compositional language understanding is handled by the image model itself.
\vspace{-4pt}
\section{Erasing contextual information from text embeddings}
\begin{figure}[t]
    \centering
    \includegraphics[width=\linewidth]{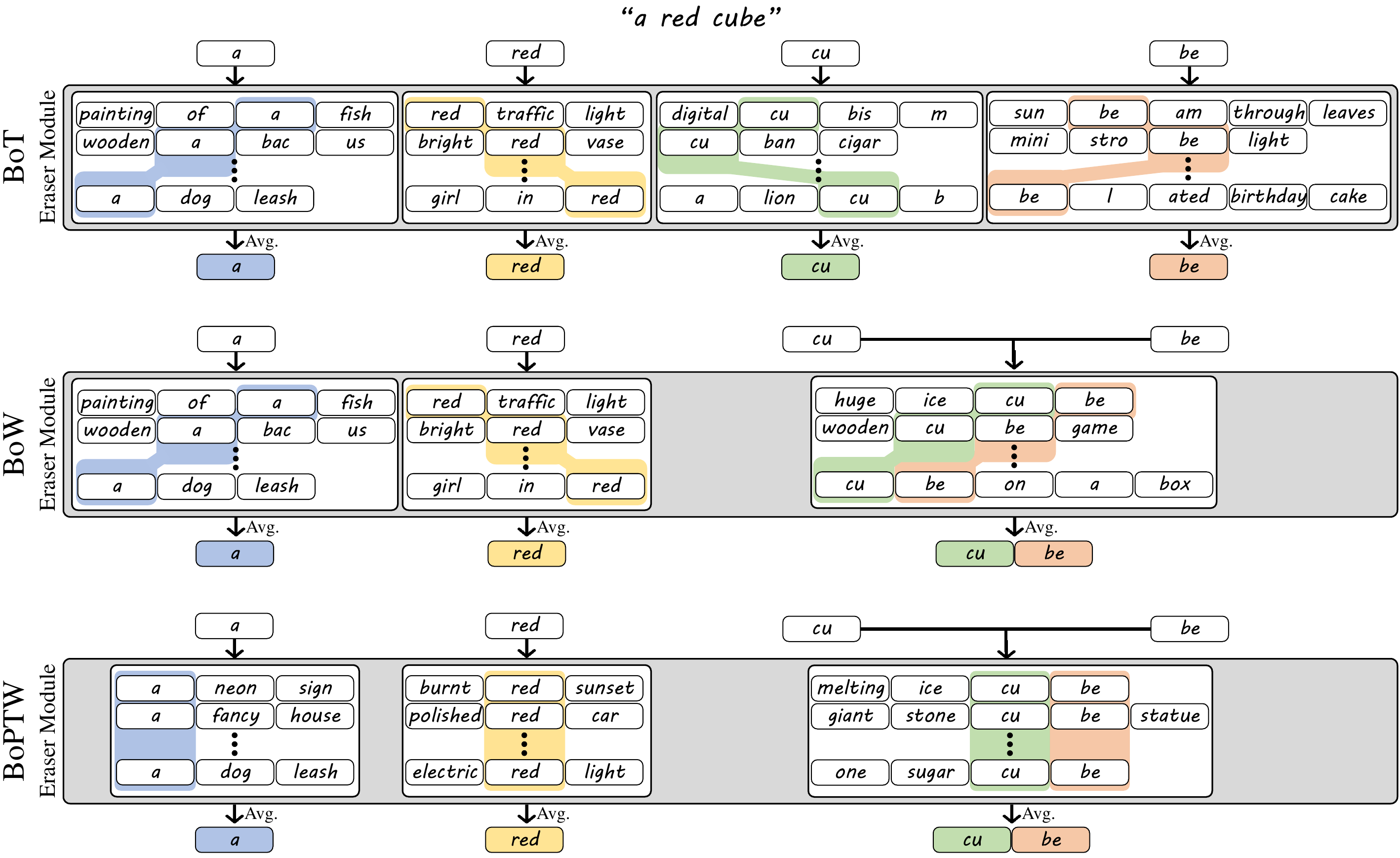}
    \caption{\textbf{Construction of contextless embedding.} To understand which types of information in the text embeddings are primarily utilized by the image model, we construct three contextless embeddings. Each begins by tokenizing the prompt (\eg ``\textit{a red cube}'') into discrete tokens by the text encoder's tokenizer (\eg ``\textit{a}'',``\textit{red}'',``\textit{cu}'',``\textit{be}''). These are processed by an eraser module which strips targeted contextual information, resulting in per-token embeddings that are concatenated to form the final embedding. 
    For the BoT embedding (top), the eraser module marginalizes across sentences where the token appears in diverse positions and contexts.
    In the BoW embedding (middle) sub-tokens of a single word (\eg ``\textit{cu}'',``\textit{be}'' for ``\textit{cube}'') are averaged exclusively across sentences where they form that word.
    Finally, the BoPTW embedding (bottom) is created by averaging only across sentences where the word appears at the same absolute position as in the original prompt.
    }
    \label{fig:erasing}
\end{figure}

\vspace{-4pt}
\label{sec:erasing}
Standard TTI models are conditioned on text embeddings. These are obtained by segmenting the user-provided text into tokens using a tokenizer, and then jointly mapping the sequence of discrete tokens into continuous embedding vectors that capture both the semantic content and the contextual structure of the prompt. The latter stage is performed by a text encoder network and we refer to the text representations at its output as \textit{full embeddings}. 
Our goal is to understand which types of information in the text embeddings are primarily utilized by the image model. To this end, we construct alternative, contextless embeddings by bypassing the standard contextualization process. We explore three alternatives, as we detail next and illustrate in ~\cref{fig:erasing}. 

\myparagraph{Bag of Tokens (BoT) embedding.}
We start by stripping off any information that may be encoded in the embedding of each token, besides its meaning as a single entity.
This is achieved as follows. For each token in the prompt, we collect a set of sentences that contain that token in various positions (see \cref{sec:sentences} for details). We pass each of these sentences through the text encoder, and average the embeddings of all appearances of the token of interest. By construction, this yields a token representation that lacks any information about the other tokens in the original prompt or about the location of the token within the prompt.
This erasure process is illustrated in the top pane of \cref{fig:erasing} for the prompt ``\textit{a red cube}''. 
After processing each token independently, we form our BoT embedding by concatenating these resulting contextless vectors and padding the remaining positions with the embedding of an empty string to reach the maximum sequence length.

\myparagraph{Bag of Words (BoW) embedding.}
BoT embeddings are inherently ambiguous. In particular, because many words are broken by the tokenizer into multiple tokens, there generally does not exist a unique bidirectional mapping between words and their constituent tokens. For example, the words ``\textit{housework}'' and ``\textit{workhouse}'' both break into the tokens ``\textit{house}'' and ``\textit{work}'', and thus cannot be disambiguated by the TTI model if only fed with the BoT embedding. In fact, multi-token words are quite prevalent; For example, $\sim32\%$ of the words in the MSCOCO-2014 prompts are split into multiple tokens by the T5 tokenizer (estimated based on a $30$K subset). To allow disambiguating such words, we introduce the BoW embedding, which refines the previous approach by preserving the cohesion of multi-token words. 
Specifically, in BoW, tokens representing complete words are processed identically to the BoT method, however the embeddings of words that split into multiple tokens, are averaged exclusively across sentences where the tokens appear as part of that specific word (see middle pane of~\cref{fig:erasing}, where the word ``\textit{cube}'' comprises the tokens ``\textit{cu}'' and ``\textit{be}'').
This captures internal word structure while still marginalizing out the surrounding prompt context. 

\myparagraph{Bag of Position-Tagged Words (BoPTW) embedding.}
While the BoW embeddings allow disambiguation between different words that comprise the same set of tokens, they are still invariant to permutations of words within the sentence. As exemplified in ~\cref{fig:teaser}, sentences comprising the same set of words can often convey different meanings (\eg ``\textit{a white box on a black box}'' vs.~``\textit{a black box on a white box}''). To allow disambiguation on the sentence level, we present the BoPTW embedding. This embedding extends the BoW method by preventing the context-erasure process from removing any positional information that may be embedded within each token. 
In~\cref{sec:position} we show that text encoders indeed embed information about the ordinal position of each token within the sequence. This is shown by observing that a token's position can be accurately inferred from its embedding. Thus, to construct the BoPTW embedding of a token, we average only over sentences where that token appears at the same position as in the prompt (in addition to belonging to the same word as in the prompt for multi-token words). This is illustrated in the bottom pane of~\cref{fig:erasing}, where \eg the embedding for the token ``\textit{red}'' is obtained by averaging only over sentences where ``\textit{red}'' appears at the second token position. We find this strategy preferable over only tagging tokens with position and not binding multi-token words (see~\cref{sec:position}).
Importantly, while the BoPTW embedding allows the TTI model to utilize both word identity and order, it still lacks any contextual information, as it does not encode semantic relationships between different words in the original prompt.
\vspace{-4pt}
\begin{figure}[t]
    \centering
    \includegraphics[width=\linewidth]{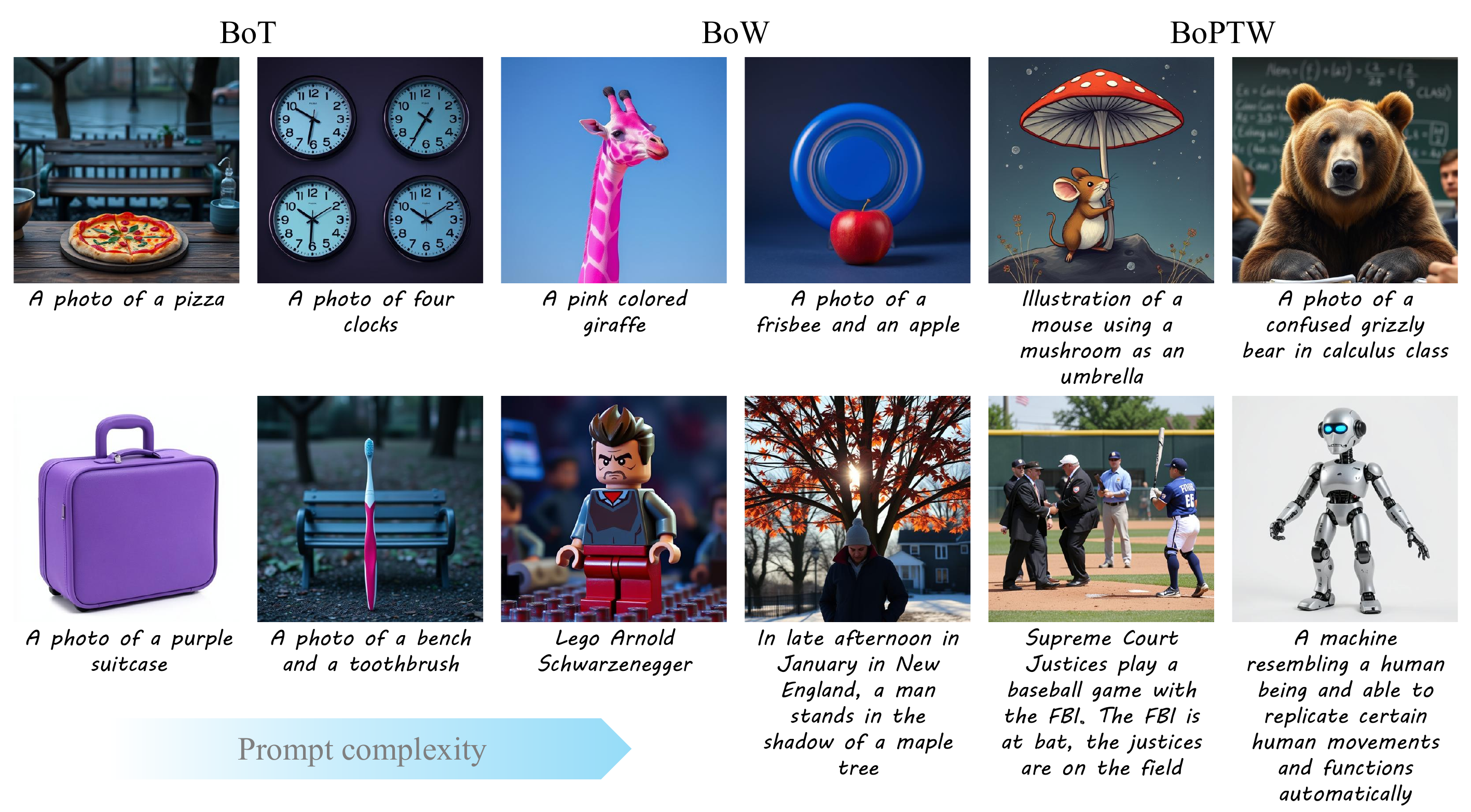}
    \caption{\textbf{Visual examples by prompt complexity}. The BoT and BoW embeddings provide the image model with sufficient information for relatively simpler cases. The BoPTW embedding can support more complex prompts. All images were generated with FLUX.1 Schnell.}
    \label{fig:variants}
\vspace{-1em}
\end{figure}

\vspace{-4pt}
\section{Results and Discussion} \label{sec:discussion}
\vspace{-4pt}
\subsection{Experimental Setup} \label{sec:exp_setup}
\vspace{-4pt}
\myparagraph{Models.} We experiment with three diffusion transformer (DiT) TTI models: SD~3~\citep{esser2024sd3}, FLUX.1 Schnell~\citep{flux2024}, and FLUX.2 Klein-4B~\citep{flux2_klein}, and generate images of size $512\times512$. The former two employ multiple text encoders. Specifically, SD~3 uses embeddings from T5-XXL, CLIP ViT-L/14, and CLIP ViT-bigG/14, along with the pooled outputs of the two CLIP encoders. Similarly, FLUX.1 uses T5-XXL embeddings together with the pooled output of CLIP ViT-L/14. FLUX.2 uses embeddings from Qwen3. 
We construct contextless embeddings for each of the text encoders used by each TTI model. To obtain contextless variants for the pooled CLIP embeddings, we apply pooling to the per-token contextless CLIP embeddings in the same manner used for the full embeddings.

\begin{figure}[t]
    \centering
    \includegraphics[width=\linewidth]{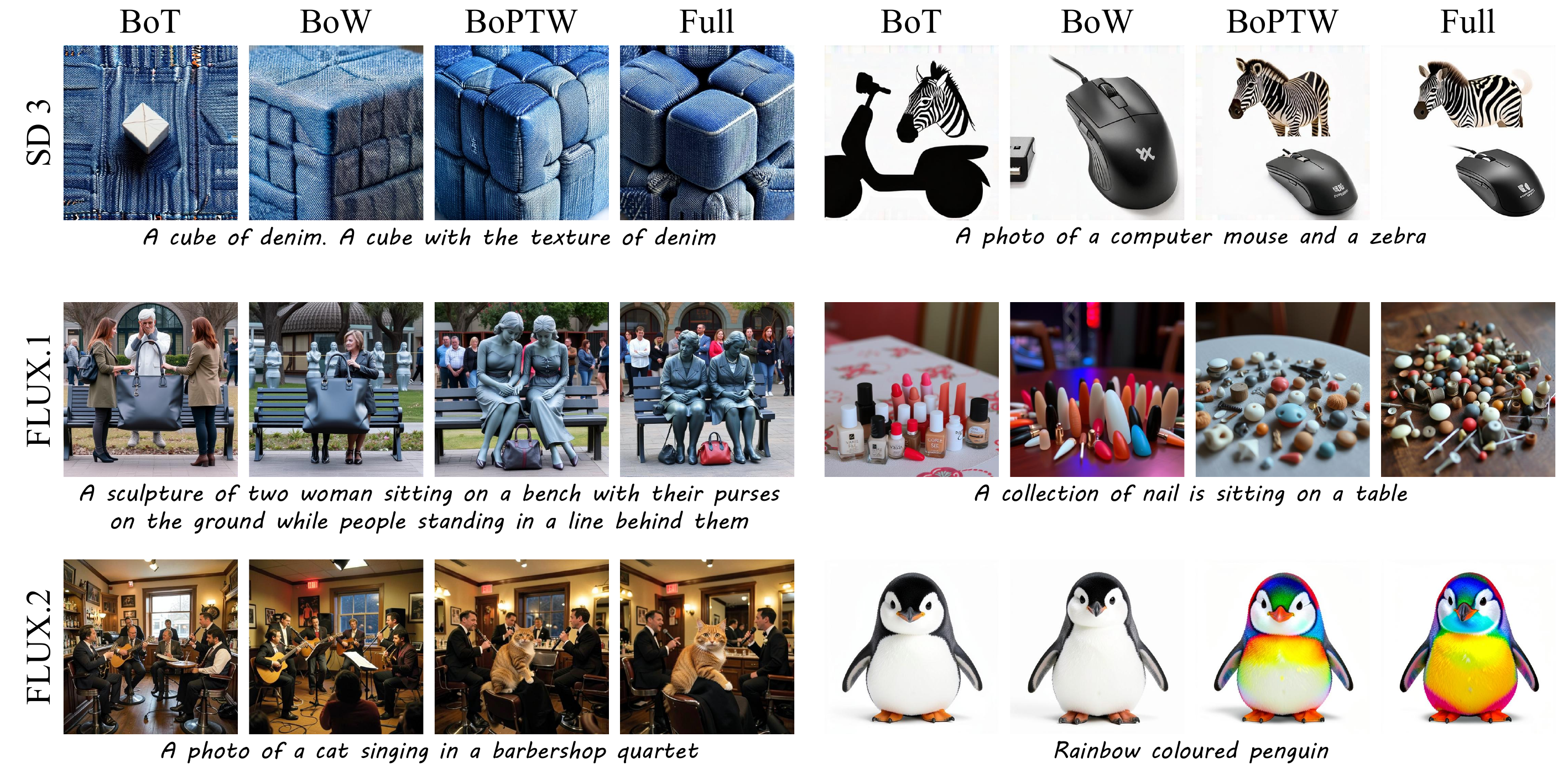}
    \caption{\textbf{Image generation with the contextless embeddings.} For complex text prompts, the BoT embeddings do not suffice for generating text-adherent images. While the BoW embeddings sometimes provide sufficient improvement, the combination of word-level tokenization with positional information provided by the BoPTW embeddings, consistently enables generating images that adhere to the prompt and are comparable in quality to those produced with the full embedding.}
    \label{fig:qualitative}
\end{figure}

\myparagraph{Construction of contextless embeddings.} As described in \cref{sec:erasing}, the construction of the contextless embedding of a token involves averaging over multiple sentences containing that token. While this process can be done on-the-fly for each new prompt, it is also possible to prepare in advance the averaged contextless representations for many of the tokens in the tokenizer's dictionary. We do this by extracting the full embeddings of all the prompts from the CC3M~\citep{changpinyo2021conceptual} and MSCOCO 2017~\citep{lin2014microsoft} training sets. For every token that appears within those datasets, we store the average of the embeddings of all its appearances in a new, ``contextless dictionary''. This is done once without respecting position (as required for the BoT and BoW embeddings) and once while respecting position (for the BoPTW embeddings). For tokens that appear less than ten times within these datasets, we use Claude-sonnet-4.5~\citep{anthropic2025claude45} to generate additional prompts containing that token so as to complete the pool to ten. For the BoW and BoPTW embeddings of tokens that are part of a word, we also use this LLM at inference time to generate sentences containing that word at a target position.

\myparagraph{Evaluation datasets.} We use prompts from three datasets for evaluation: DrawBench~\citep{saharia2022photorealistic}, GenEval~\citep{ghosh2023geneval}, and a $30$K subset of the MSCOCO-2014 validation set~\citep{lin2014microsoft}. Together, these form a curated collection of complex, human-authored prompts designed to rigorously assess fine-grained text-image alignment. DrawBench and GenEval partition the prompts into categories, including attribute binding, spatial reasoning, counting, color-object consistency, and multi-object compositionality, which we utilize to obtain a per-task dissection of the successes and failures of the contextless embeddings. We exclude from DrawBench the categories of misspellings and rare words, on which the examined TTI models often fail even with the full embeddings.
For DrawBench and GenEval we generate five images per prompt, and for MSCOCO-2014 we generate only one due to this dataset's scale.

\myparagraph{Evaluation metrics.}
We use the vision-language model (VLM) Gemma-3~\citep{gemmateam2025gemma3technicalreport} to compare between images generated from the full embeddings and those produced using our contextless embeddings. Gemma is used as an automated evaluator in a blind three-way comparison setting, tasking it to choose one of the following options for each prompt: ($i$) the image generated from the contextless embedding is preferred over that from the full embedding, ($ii$) the image generated from the full embedding is preferred over that from the contextless embedding, or ($iii$) no clear preference among the two. \Cref{sec:vlm_eval} provides the precise instructions to the VLM. It should be noted that the contextless embeddings are not a-priori expected to outperform the full embeddings; we are merely interested in analyzing when they do not lead to worse results. Therefore, while we report the proportions for all three options, we mostly care about our embeddings' \textit{non-inferiority rate}, which is the proportion of times they were not ranked as worse than the full embeddings (cases ($i$) and ($iii$)). Yet, in practice we find that the contextless embeddings are preferable in a surprisingly non-negligible proportion of the cases. Therefore, we regard the non-inferiority rate of the full embeddings (cases ($ii$) and ($iii$)) as a baseline for comparison. 
In \cref{subsec:metrics} we complement the VLM analysis with other metrics for quantifying the quality of the generated images and their adherence to the prompt, including CLIP score, FID~\citep{heusel2017gans}, and KID~\citep{binkowski2018demystifying}. These show the same trends as the VLM evaluation. 
In \Cref{sec:vlm_agreement}, we additionally report results obtained with Qwen3~\citep{yang2025qwen3}, along with its agreement with the Gemma-3 evaluator.

\vspace{-4pt}
\subsection{Results}
\vspace{-4pt}
\label{sec:results}
\myparagraph{Qualitative results.}
\Cref{fig:teaser} presents images generated with the BoPTW embedding. The two prompts in each pair comprise the same set of words, only in a different order. Although the BoPTW embedding of each word in the prompt lacks any context from the other words, all the examined models succeed in disambiguating the prompt meanings from those embeddings. \Cref{fig:variants} depicts successful generations with all three contextless embedding types using SD~3. It can be observed that for simple prompts, even the BoT embeddings succeed. Generations for more complex prompts are enabled by the BoW and BoPTW embeddings. \Cref{fig:qualitative} illustrates how the generation improves when switching from BoT to BoW, and to BoPTW. 
For instance, in the prompt ``\textit{A cube of denim. A cube with the texture of denim},'' the multi-token word ``\textit{cube}'' becomes better resolved when switching from BoT to BoW.
In some cases, the information regarding the absolute position of the word is what leads to a significant improvement. For example, for the prompt ``\textit{A sculpture of two women sitting on a bench with their purses on the ground while people standing in a line behind them},'' capturing the full structure of the scene requires both word-level and positional information. Only when both are incorporated (using the BoPTW embedding) does the image model successfully infer the intended relationships and produce an image comparable to that obtained with the full embedding.

\begin{figure}
    \centering
    \includegraphics[width=\linewidth]{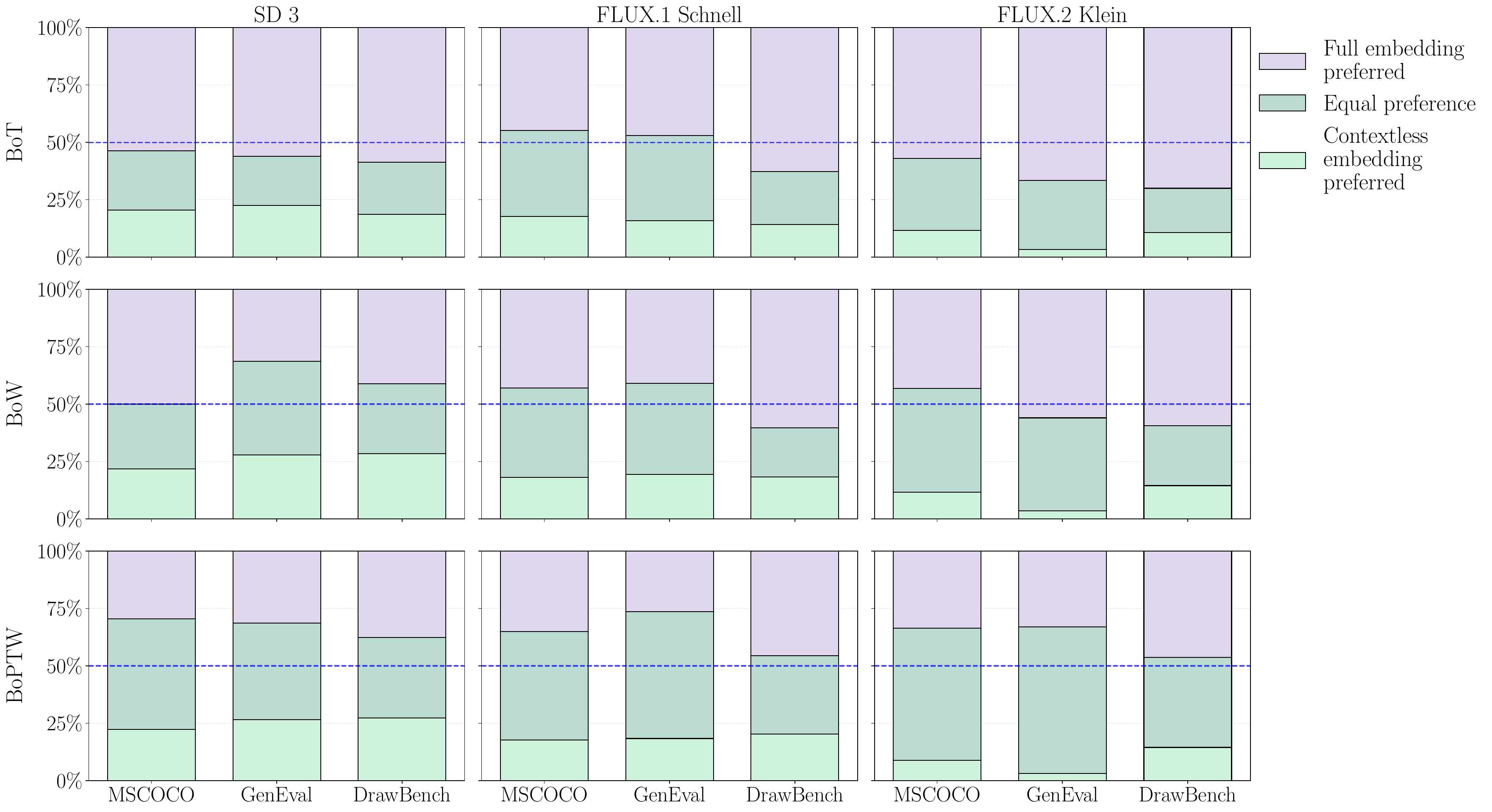}
    \caption{\textbf{Text alignment comparison.} Image pairs generated from full versus contextless embeddings are compared using Gemma as an automated evaluator. 
    Notably, the BoPTW embedding (bottom row) achieve a non-inferiority rate of at least $65\%$ with respect to the full embedding (the combination of the two greenish areas) for most benchmarks and models. This is while the non-inferiority rate of the full-embedding with respect to  BoPTW (the combination of the purple and dark green areas) is typically only $70\%-90\%$ for most models and datasets.
    }
    \label{fig:erasing_results}
\end{figure}

\myparagraph{Quantitative comparisons.}  \Cref{fig:erasing_results} presents the VLM responses for all models and datasets. 
Surprisingly, we observe that even the simplest \textit{BoT embeddings} (top row) already achieve a non-inferiority score that exceeds $40\%$ for most experimental settings. This demonstrates that TTI models are often able to correctly decipher the meanings of prompts only from token identities, without even requiring knowledge on their ordinal positions within the prompt. Nevertheless, the non-inferiority rate of the BoT embeddings usually remains below $50\%$, implying that there are many prompts that do require richer representations for faithfully capturing semantics. But how much richer?
It turns out that the \textit{BoW embeddings} (second row), which are only slightly richer, already yield a notable improvement, raising the non-inferiority rate above $50\%$ for most datasets and models. This shows that TTI models often only need the context embedded within tokens to determine which sub-word tokens make up a single word in the prompt.
Still, a noticeable discrepancy persists for FLUX.2, suggesting that this model remains sensitive to the loss of richer contextual structure in the embedding space. 
The \textit{BoPTW embeddings} (third row), which additionally contain absolute word-level positional information, lead to a non-inferiority rate that reaches $65\%$ for most models and datasets, coming close to the non-inferiority rate of the full embedding, which is between $70\%-90\%$ for most models and datasets.
This suggests that the combination of word-level tokenization and absolute token position is commonly sufficient for the image model to reconstruct the necessary contextual information internally. \Cref{tab:clip} in the appendix reports CLIP-based prompt adherence scores, demonstrating that BoPTW achieves scores that are consistently close to those of the full embeddings. The FID and KID scores, reported in Tables~\ref{tab:fid_inc} and \ref{tab:kid_inc} in the appendix, indicate that the image quality is comparable across all embedding types.

\begin{figure}[t]
    \centering
    \includegraphics[width=1\linewidth]{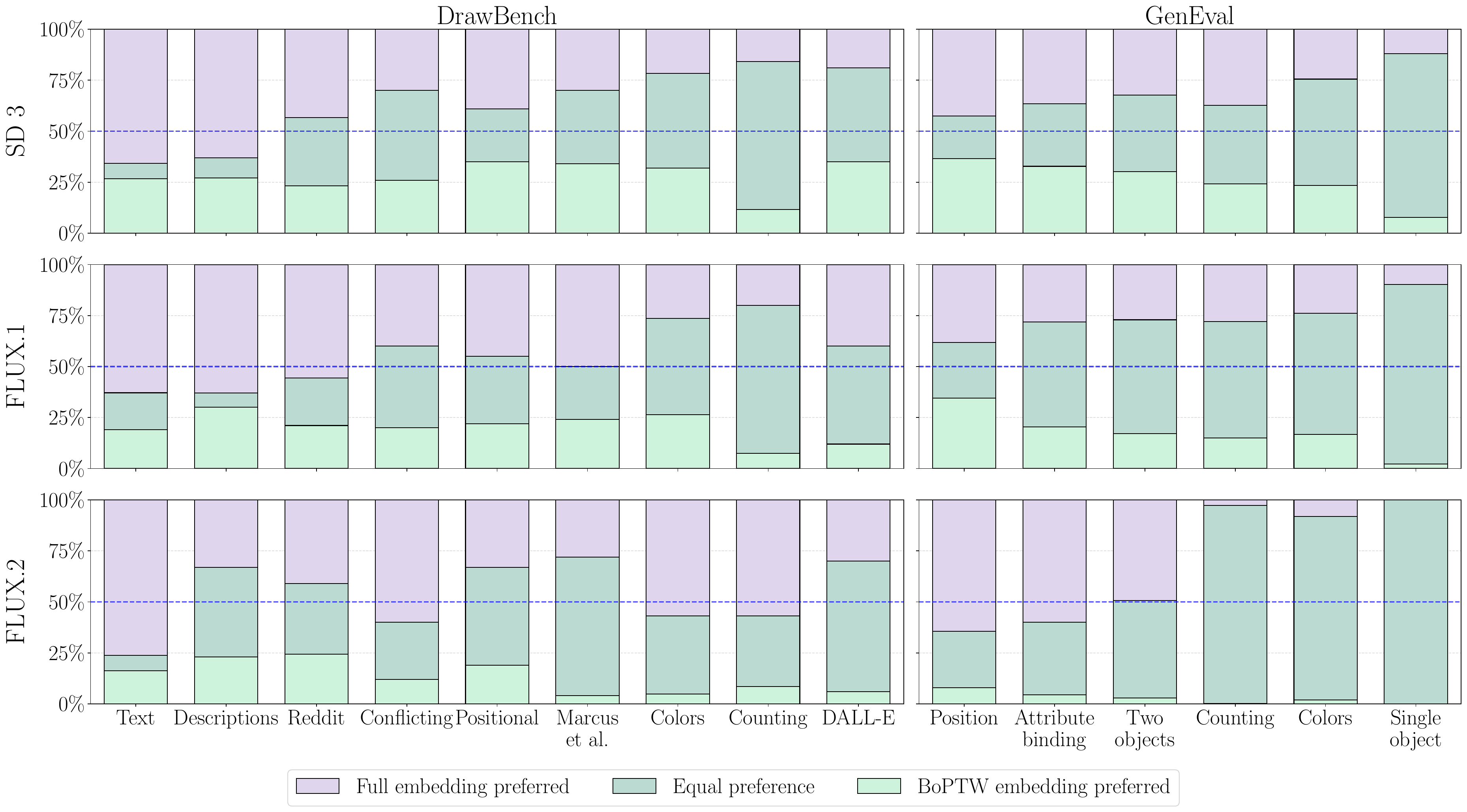}
    \caption{\textbf{Text alignment across categories.} Breakdown on VLM responses by category on images generated with the BoPTW embedding, for the DrawBench and GenEval benchmarks. Results are sorted by the mean non-inferiority rate across all evaluated models to highlight which categories are most resilient to the removal of full prompt context. For the GenEval dataset, we further report in~\cref{tab:geneval} the task-specific scores according to the evaluation framework of that dataset.}
    \label{fig:categories_bars}
\end{figure}

\myparagraph{Breakdown according to prompt categories.} 
In \cref{fig:categories_bars}, we present the VLM responses for the BoPTW embedding for each of the categories of the DrawBench and GenEval datasets separately. 
The categories are sorted by average non-inferiority. \Cref{fig:categories} visualizes the most and least successful ones. As \cref{fig:categories_bars} illustrates, non-inferiority rates exceed $50\%$ for most categories, and some reach much higher values.
For example, on the ``Single object'' category in GenEval, the non-inferiority rates of BoPTW are $88\%$, $90\%$, and $100\%$ with SD~3, FLUX.1, and FLUX.2, respectively (see first row of \cref{fig:categories}).
On the other hand, some categories are challenging for generation with contextless embeddings. 
For example, on the ``Text'' category in DrawBench, the rates are only $27\%$, $37\%$, and $24\%$ with SD~3, FLUX.1, and FLUX.2, respectively (see
third row of \cref{fig:categories}). 
This is related to how the presence of quotation marks alters a word's embedding. Specifically, a word enclosed in quotation marks (e.g., ``cat'') receives a different embedding than the same word without quotation marks (cat).
See more visual examples in \cref{sec:additional_geneval} and \cref{sec:additional_DrawBench}.
\Cref{tab:geneval} reports a breakdown of the results on the GenEval benchmark using the original paper's evaluation protocol~\citep{ghosh2023geneval}. In those metrics, BoPTW performs on par with the full embeddings, and in several cases even surpasses them.

\myparagraph{Long prompts}
We evaluate the contextless embeddings also on the DPG-Benchmark~\citep{hu2024ella}, which is designed to assess TTI performance on long, complex, and highly detailed prompts.
The benchmark consists of 1,065 prompts, with an average length of 408 words per prompt. Following the experimental setup described in~\Cref{sec:exp_details}, we generated five images per prompt using both the contextless and full embeddings and evaluated the resulting images with Gemma-3.  
BoPTW achieved non-inferiority rates of $47.2\%$, $57.0\%$, and $41.5\%$ for SD~3, FLUX.1, and FLUX.2, respectively. See more results in~\Cref{ssection:Evaluation on DPG benchmark}.

\myparagraph{User study}
To validate our automatic evaluation, we conducted a user study comparing the text adherence of images generated with BoPTW embeddings versus full embeddings. On subsets of both DrawBench and GenEval, human evaluators found no significant advantage for either method in the majority of cases ($63\%$ and $60\%$, respectively), corroborating our VLM-based non-inferiority results. 
Furthermore, the VLM–user agreement rates closely approach the user–user agreement rates, suggesting that both Gemma-3 and Qwen3 provide a reliable proxy for human preferences. Full details and a complete quantitative breakdown of this study are provided in \Cref{ssection:User Study}.

\vspace{-4pt}
\subsection{Discussion}
\vspace{-4pt}
While Imagen~\citep{saharia2022photorealistic} highlighted the necessity of dedicated text encoders for U-Net-based architectures, we investigate whether these older models rely more heavily on contextualized text embeddings than modern DiTs. Evaluating the legacy models SDXL~\citep{podell2023sdxl} and SD~2.1~\citep{rombach2022ldm} on DrawBench under the BoPTW setting reveals that U-Net architectures completely fail with contextless embeddings. Specifically, as shown in \cref{fig:non_dit}, non-inferiority rates plummet to $0.2\%$ for SD~2.1 and $4\%$ for SDXL. This stark contrast suggests a fundamental shift in linguistic processing: DiT models are robust enough to interpret complex language directly, whereas U-Nets depend heavily on the text encoder. Because text encoders are optimized for text-only tasks, their embeddings encode contextual nuances often irrelevant to image generation. DiTs successfully filter out this irrelevant variability, indicating they achieve a more optimal division of linguistic understanding between the text encoder and the image model. 
To validate this insight, we follow \citep{wang2026circuit}, and compare DiT and U-Net on a synthetic benchmark using the same frozen T5 encoder. Under BoPTW versus full embeddings, DiT achieved a $67.56\%$ non-inferiority rate compared to U-Net's $26.27\%$, demonstrating that DiT successfully grounds linguistic structure while U-Net fails. 
Also, as we show in \Cref{fig:mech}, our contextless BoPTW condition produces initially diffuse attention that dynamically converges over time. This proves the image model internally decodes linguistic structures.
See \Cref{sec:Discussion_app} for more details.

\vspace{-4pt}
\section{Conclusion}
\vspace{-4pt}
We challenged the prevailing assumption that TTI models rely on the rich, contextualized embeddings provided by complex text encoders.
By demonstrating that a Bag-of-Position-Tagged-Words embedding is sufficient to maintain text fidelity and visual quality, we revealed that transformer based diffusion models primarily rely on individual word meanings and their relative order rather than full prompt context.
Broadening this representation from words to multi-word idioms may further improve these results.
Our observations suggest that linguistic decoding is performed mostly by the image model, rather than relying on the text encoder. 
This shift in understanding opens new avenues for training TTI models on simpler word-level embeddings augmented with explicit position tags rather than large text encoders, simplifying the interface between language and vision.

\subsection*{AI use statement}
In this work, generative AI tools were utilized exclusively for three purposes: employing VLMs as automated judges for image generation (\Cref{sec:exp_details}), using Claude-sonnet-4.5 to generate synthetic prompts for the construction of contextless embeddings, and refining the manuscript's phrasing. 
No other components of this research relied on AI assistance. We take full responsibility for the entirety of this work, including all text, claims, and artifacts produced with the aid of generative AI.

\bibliography{main}
\bibliographystyle{iclr2027_conference}

\appendix
\section*{Appendix}
\FloatBarrier
\section{Additional Results}
\subsection{Encoding positional information in the text embedding}
\label{sec:position}

\begin{figure}[ht]
    \centering
    \begin{subfigure}{0.3\textwidth}
        \centering
        \includegraphics[width=\textwidth]{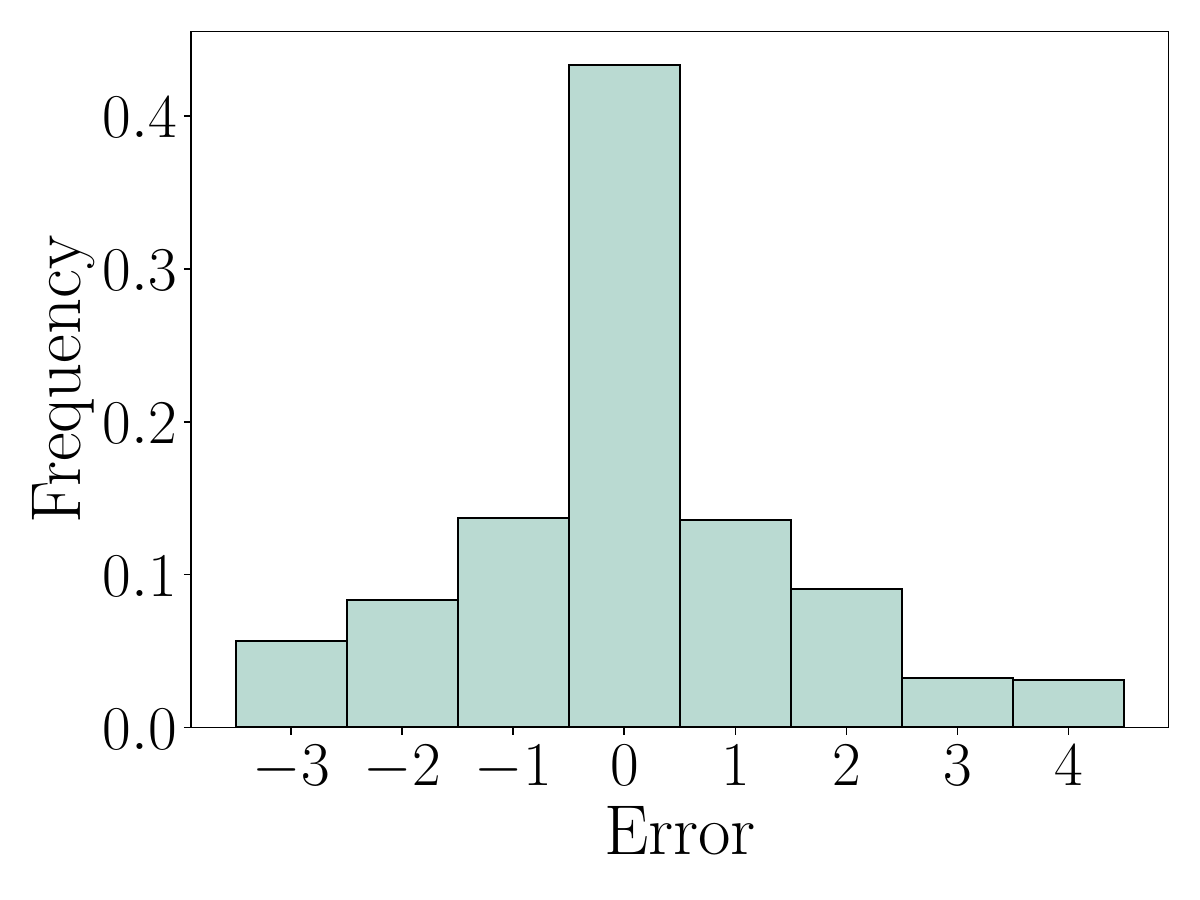}
        \caption{CLIP}
    \end{subfigure}
    \hfill
    \begin{subfigure}{0.3\textwidth}
        \centering
        \includegraphics[width=\textwidth]{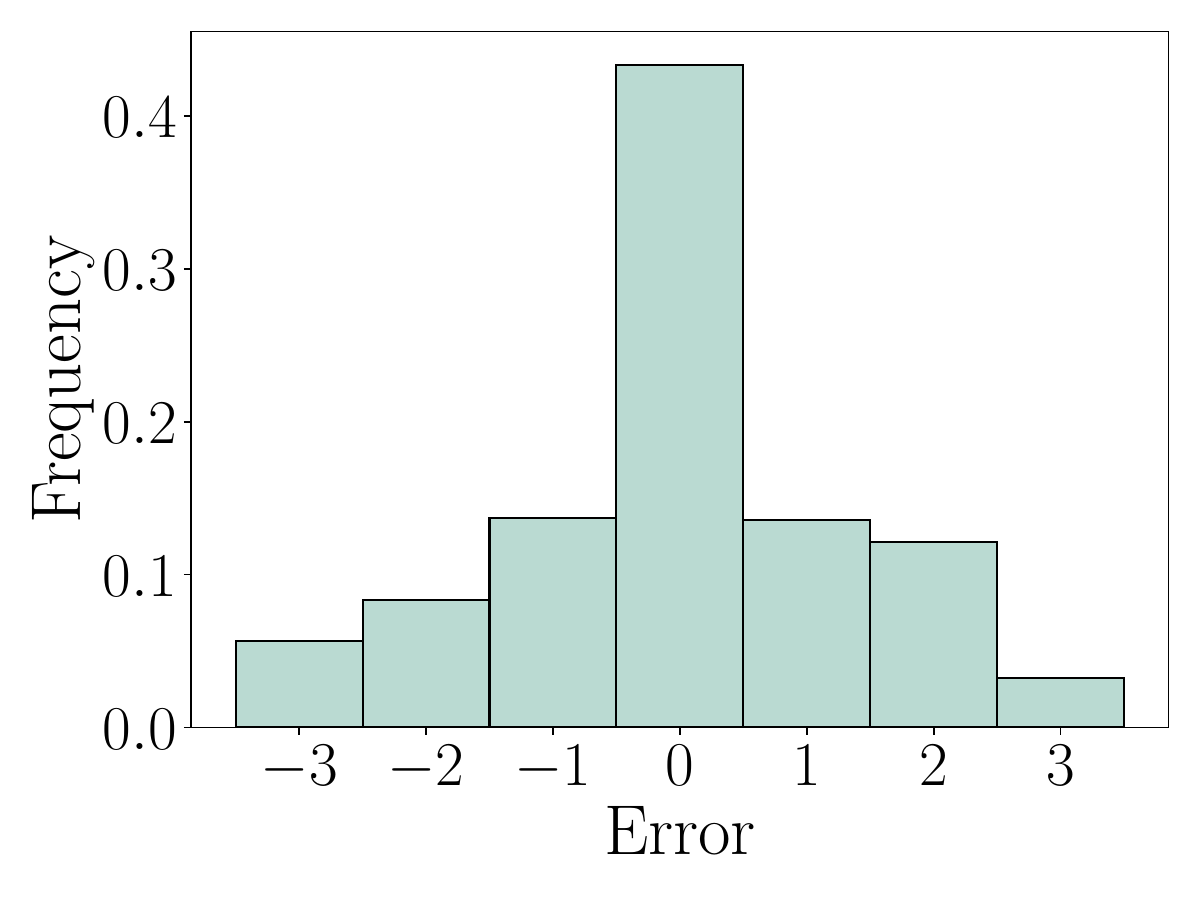}
        \caption{T5}
    \end{subfigure}
    \hfill
    \begin{subfigure}{0.3\textwidth}
        \centering
        \includegraphics[width=\textwidth]{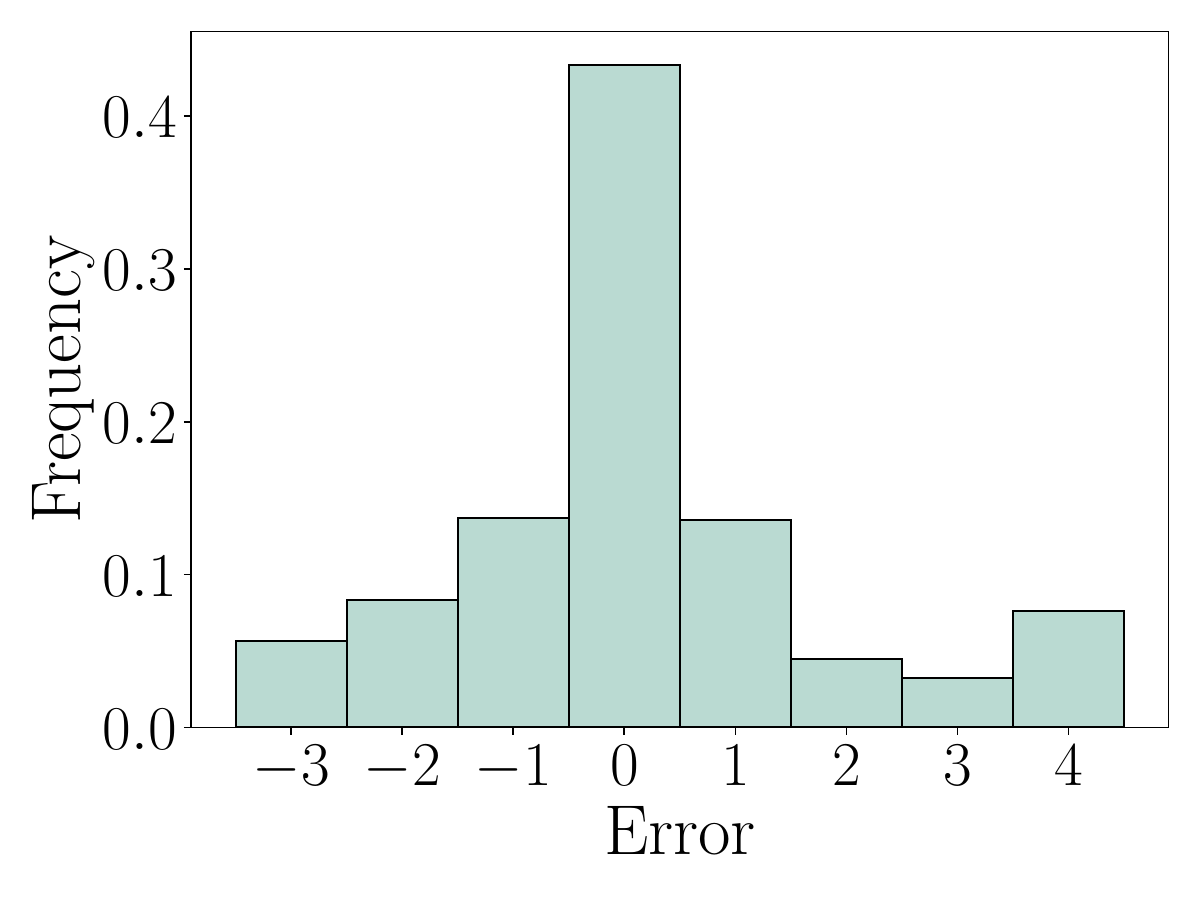}
        \caption{Qwen}
    \end{subfigure}
    \caption{\textbf{Positional information in token embeddings.} Distribution of positional prediction errors across three text encoders. All show that majority of tokens are classified with no error, confirming that text encoders encode information about the ordinal position of each token in the sentence.}
    \label{fig:predicting_position}
\end{figure}

To empirically evaluate the positional knowledge encoded in token representations, we sampled $230$K tokens from the MSCOCO dataset that appear in different positions in their original sentences. 
For each sampled token, we extract its embedding from the text encoder and compute its cosine similarity to a dictionary of token-specific reference embeddings constructed for BoPTW embeddings.
Each entry in this dictionary represents the average embedding for a given token at a specific ordinal position. We then predicted the token’s position by selecting the index of the dictionary entry with the highest similarity.
The error is calculated as the distance between this predicted index and the ground-truth position. As shown in \cref{fig:predicting_position}, the error distribution reveals that across all three evaluated text encoders, the vast majority of tokens were localized with zero error, confirming that absolute positional information exists.

\begin{figure}[h]
    \centering
    \begin{subfigure}{\textwidth}
        \centering
        \includegraphics[width=\textwidth]{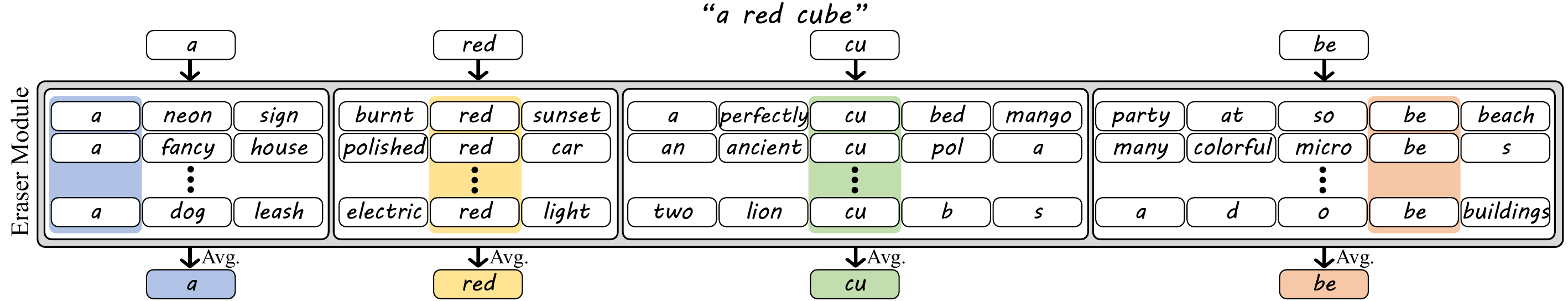}
        \label{fig:boptt_erasing}
    \end{subfigure}
    \begin{subfigure}[t]{\textwidth}
        \centering
        \includegraphics[width=\textwidth]{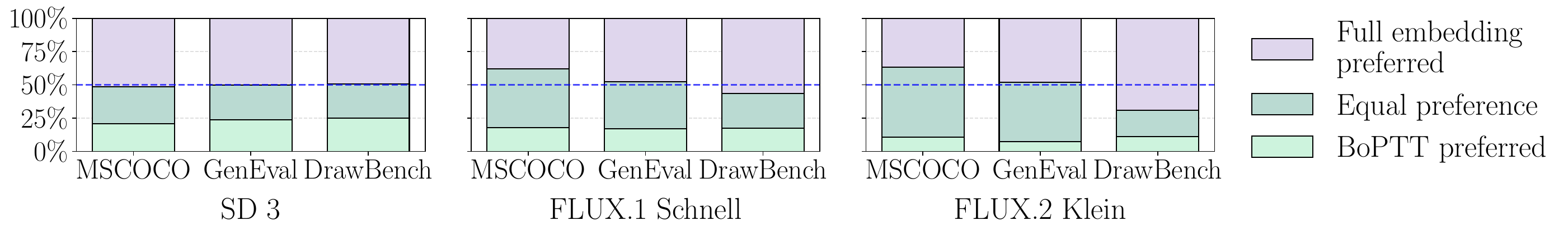}
        \label{fig:boptt_results}
    \end{subfigure}
    
    \caption{\textbf{Bag of Position-Tagged Tokens.} Top pane illustrates the construction of BoPTT embeddings, where each token is averaged only across sentences where it appears in the same position. Bottom pane presents text alignment performance  across three models and datasets.}
    \label{fig:boptt}
\end{figure}

\myparagraph{Bag of Position-Tagged Tokens (BoPTT) embedding.} 
As an extension of the token-level analysis, we evaluate the BoPTT embedding. Similarly to how BoPTW expands the BOW embedding, this embedding expands BoT embeddings by introducing spatial structure over the tokens. This is achieved by averaging the representation of each token across a set of sentences where the token appears at the same ordinal position as in the target prompt, illustrated in the top pane of~\cref{fig:boptt}.
The text adherence, reported in the bottom pane of~\cref{fig:boptt}, shows improvement over the BoT embedding, bringing the non-inferiority rate of most settings to $50\%$.

\clearpage
\subsection{Metrics}
\label{subsec:metrics}
In \cref{tab:clip}, we report CLIP-based prompt adherence scores comparing images generated with the full embedding to those generated using our contextless embeddings. 
Across all contextless variants, CLIP scores are comparable to those obtained with full embeddings and exhibit trends consistent with VLM evaluation presented in Sec.~\ref{sec:results}.
Tables~\ref{tab:fid_inc} and \ref{tab:kid_inc} report FID and KID scores respectively, both computed using the \texttt{inception-v3-compat} feature extractor. Across both metrics, contextless embeddings do not degrade the quality of the generated images.

\begin{table}[ht]
    \caption{CLIP-based prompt adherence scores. Higher scores correspond to better results.}
    \label{tab:clip}
    \small
    \begin{center}
    \begin{tabular}{llllllllll}
    \textbf{Benchmark} & \multicolumn{3}{c}{\textbf{DrawBench}} & \multicolumn{3}{c}{\textbf{GenEval}} & \multicolumn{3}{c}{\textbf{MSCOCO}} \\
    \textbf{Model} & FLUX.1 & FLUX.2 & SD 3 & FLUX.1 & FLUX.2 & SD 3 & FLUX.1 & FLUX.2 & SD 3 \\
    \textbf{Embedding} &  &  &  &  &  &  &  &  &  \\
    \hline \\
    Full & 33.5 & 33.8 & 33.3 & 33.7 & 34.6 & 33.4 & 31.8 & 32.0 & 31.5 \\
    BoPTW & 32.3 & 31.8 & 32.7 & 33.6 & 32.7 & 33.4 & 31.6 & 31.2 & 31.4 \\
    BoW & 30.9 & 31.4 & 32.9 & 32.8 & 30.5 & 33.4 & 31.4 & 30.9 & 31.0 \\
    BoPTT & 31.0 & 30.1 & 31.9 & 32.1 & 31.8 & 31.7 & 31.5 & 31.0 & 30.9 \\
    BoT & 30.5 & 30.1 & 30.6 & 32.1 & 28.9 & 31.7 & 31.3 & 30.1 & 30.7 \\
    \end{tabular}
    \end{center}
\end{table}

\begin{table}[ht]
    \caption{FID scores across text embeddings on MSCOCO. Lower scores correspond to better results.}
    \label{tab:fid_inc}
    \small
    \begin{center}
    \begin{tabular}{llll}
    \textbf{Model} & \textbf{FLUX.1} & \textbf{FLUX.2} & \textbf{SD~3} \\
    \textbf{Embedding} &  &  &  \\
    \hline \\
    Full & 25.6 & 27.2 & 26.2 \\
    BoPTW & 27.1 & 27.2 & 26.9 \\
    BoW & 27.6 & 26.8 & 26.2 \\
    BoPTT & 27.2 & 27.5 & 26.1 \\
    BoT & 27.6 & 26.9 & 27.8 \\
    \end{tabular}
    \end{center}
\end{table}

\begin{table}[ht]
    \caption{KID scores across text embeddings on MSCOCO. Lower scores correspond to better results.}
    \label{tab:kid_inc}
    \small
    \begin{center}
    \begin{tabular}{llll}
    \textbf{Model} & \textbf{FLUX.1} & \textbf{FLUX.2} & \textbf{SD~3} \\
    \textbf{Embedding} &  &  &  \\
    \hline \\
    Full & 0.0100 ± 0.0011 & 0.0136 ± 0.0015 & 0.0099 ± 0.0010 \\
    BoPTW & 0.0103 ± 0.0010 & 0.0134 ± 0.0013 & 0.0100 ± 0.0009 \\
    BoW & 0.0103 ± 0.0010 & 0.0132 ± 0.0014 & 0.0087 ± 0.0006 \\
    BoPTT & 0.0103 ± 0.0010 & 0.0134 ± 0.0013 & 0.0086 ± 0.0006 \\
    BoT & 0.0103 ± 0.0009 & 0.0126 ± 0.0013 & 0.0092 ± 0.0006 \\
    \end{tabular}
    \end{center}
\end{table}

\subsection{Categories breakdown}
\label{sec:Categories_breakdown}

\begin{figure}[t]
    \centering
    \includegraphics[width=\linewidth]{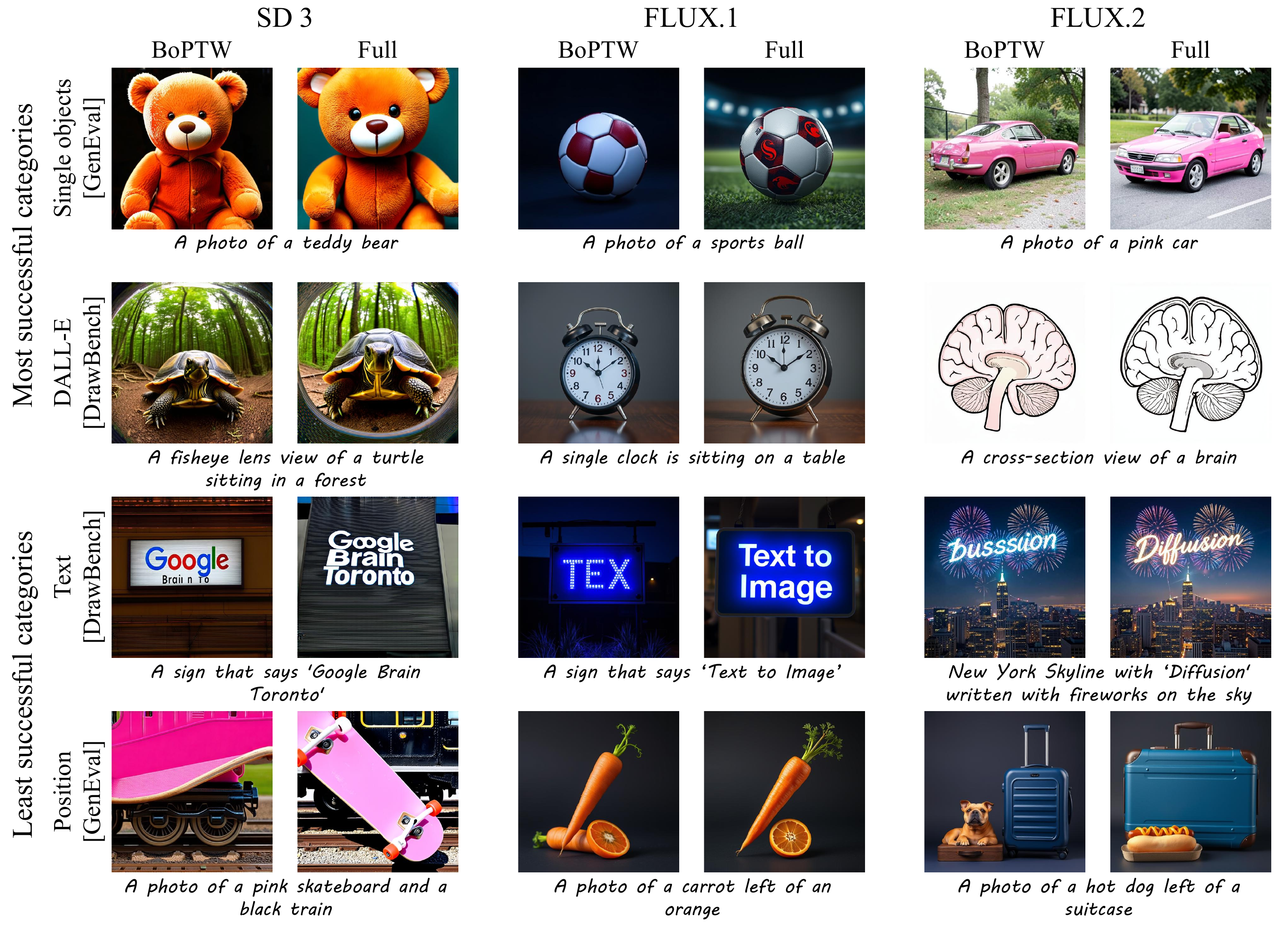}
    \caption{\textbf{Most and least successful categories.} 
    The top and bottom pairs of rows show visual examples from the two most successful and two least successful categories, respectively, in the DrawBench and GenEval datasets. Each example compares the image generated from the BoPTW embedding to that generated from the full embedding.
    }
    \label{fig:categories}
\end{figure}

\subsection{Additional results on the GenEval dataset}
\label{sec:additional_geneval}
\begin{figure}[ht]
    \centering
    \includegraphics[width=0.77\linewidth]{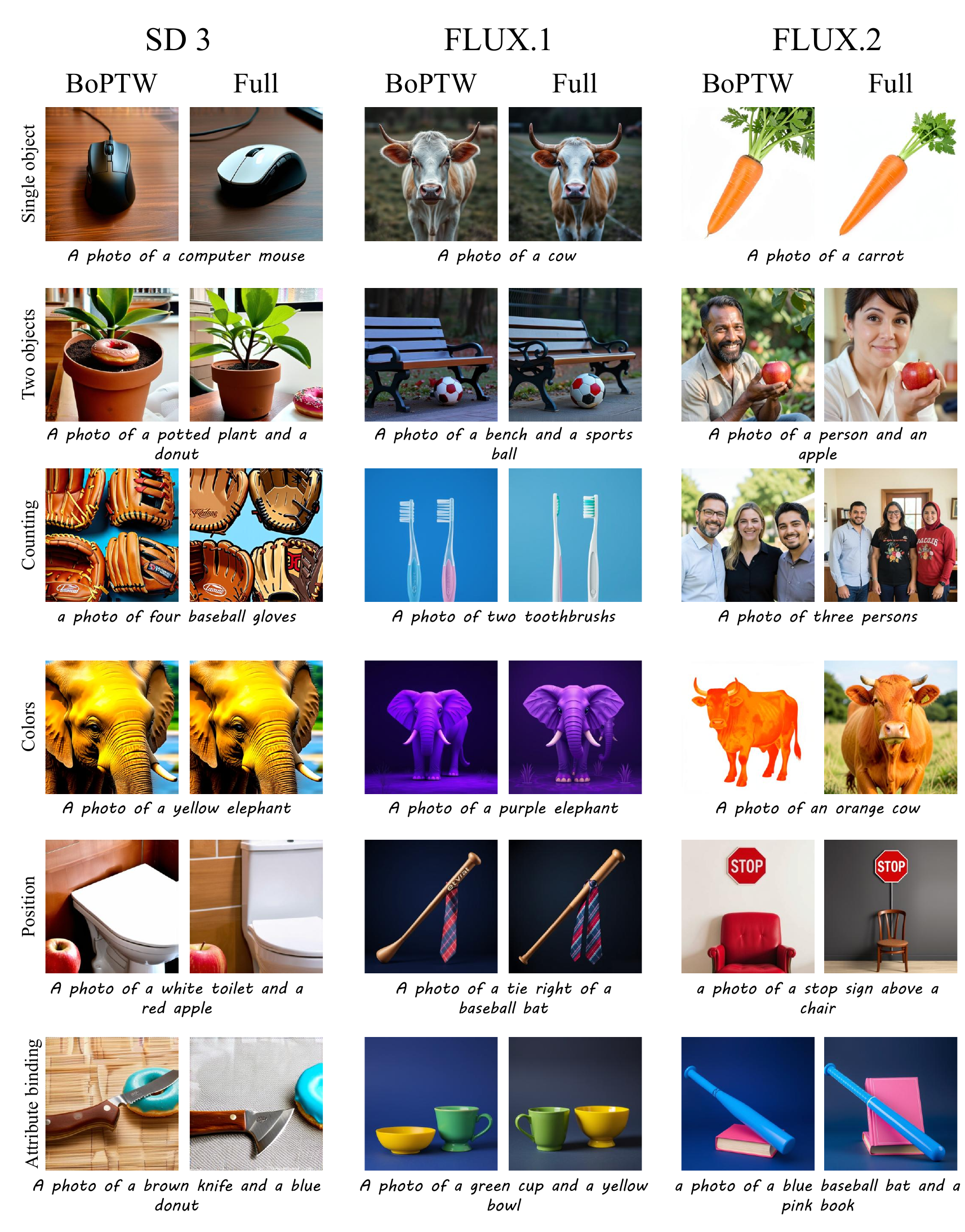}
    \caption{\textbf{Successful prompts from the GenEval categories.} We present prompts from all GenEval categories in which the BoPTW embedding was sufficient for the image model to adhere to the prompt.}
    \label{fig:geneval_supp_good}
\end{figure}

\begin{figure}
    \centering
    \includegraphics[width=0.77\linewidth]{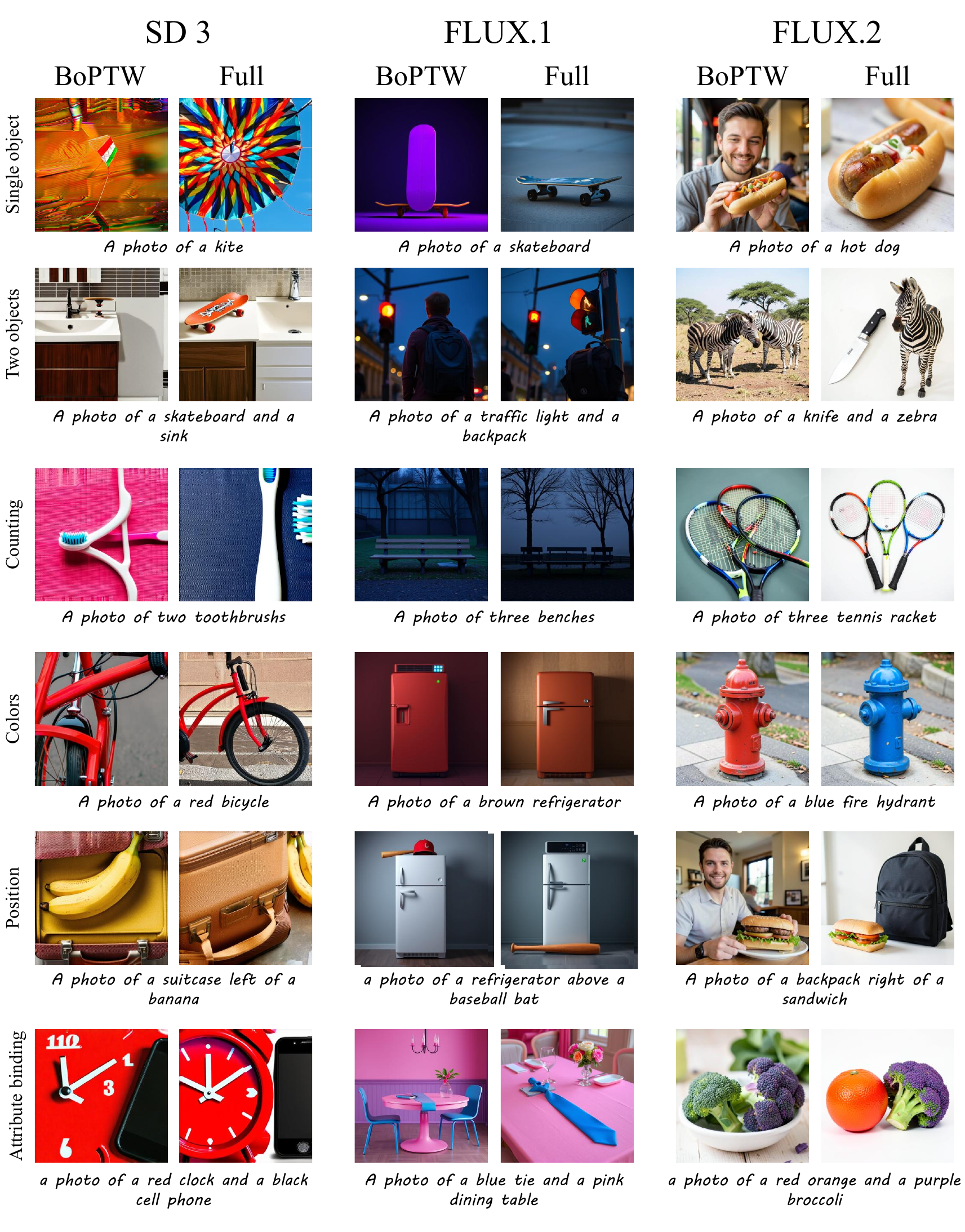}
    \caption{\textbf{Unsuccessful prompts from the GenEval categories.} We present prompts from all GenEval categories in which the BoPTW embedding wasn't sufficient for the image model to adhere to the prompt. }
    \label{fig:geneval_supp_bad}
\end{figure}

\Cref{tab:geneval} reports the GenEval scores across categories as well as overall performance, following the evaluation framework of the original paper. 
BoPTW-conditioned generations achieve comparable performance to full embeddings, occasionally surpassing the baseline in specific categories.
In addition, in~\cref{fig:geneval_supp_good} we present examples from each category of the GenEval dataset and for each of the tested models, illustrating cases in which contextless embeddings are sufficient to guide the image model to generate outputs that faithfully adhere to the text, as well as cases where they are not.
In~\cref{fig:geneval_supp_bad} we show visual examples and prompts for which the contextless embeddings are not sufficient for the image model to generate correct images.

\begin{table}
    \caption{GenEval evaluation scores for each category. Higher scores are better.}
    \label{tab:results}
    \small
    \setlength{\tabcolsep}{4pt}
    \begin{tabular}{llp{1cm}p{1cm}p{1cm}p{1cm}p{1cm}p{1.5cm}p{1cm}}
    \textbf{Model} & \textbf{Embedding} 
    & \textbf{Single\newline Object} 
    & \textbf{Two\newline Objects} 
    & \textbf{Count} 
    & \textbf{Color} 
    & \textbf{Position} 
    & \textbf{Attribute\newline Binding} 
    & \textbf{Overall} \\
    \hline \\
    
    \multirow{2}{*}{SD 3}
        & Full  & 98.0 & 82.6 & 54.5 & 81.2 & 27.8 & 52.0 & 66.0  \\
        & BoPTW           & 99.0 & 77.5 & 53.2 & 80.4 & 23.8 & 49.8 & 63.9 \\
    \\
    \multirow{2}{*}{FLUX.1}
        & Full  & 99.5 & 90.4 & 67.9 & 78.3 & 31.6 & 54.8 & 70.4  \\
        & BoPTW           & 98.7 & 88.8 & 66.2 & 79.2 & 36.3 & 55.1 & 70.7 \\
    \\
    \multirow{2}{*}{FLUX.2}
        & Full  & 99.7 & 92.7 & 83.5 & 88.7 & 62.4 & 66.8 & 82.3  \\
        & BoPTW           & 99.7 & 58.9 & 80.5 & 83.6 & 55.8 & 49.5 & 71.3 \\
    \end{tabular}
    \label{tab:geneval}
\end{table}

\clearpage
\subsection{Additional results on the DrawBench dataset}
\label{sec:additional_DrawBench}
\begin{figure}[ht]
    \centering
    \includegraphics[width=\linewidth]{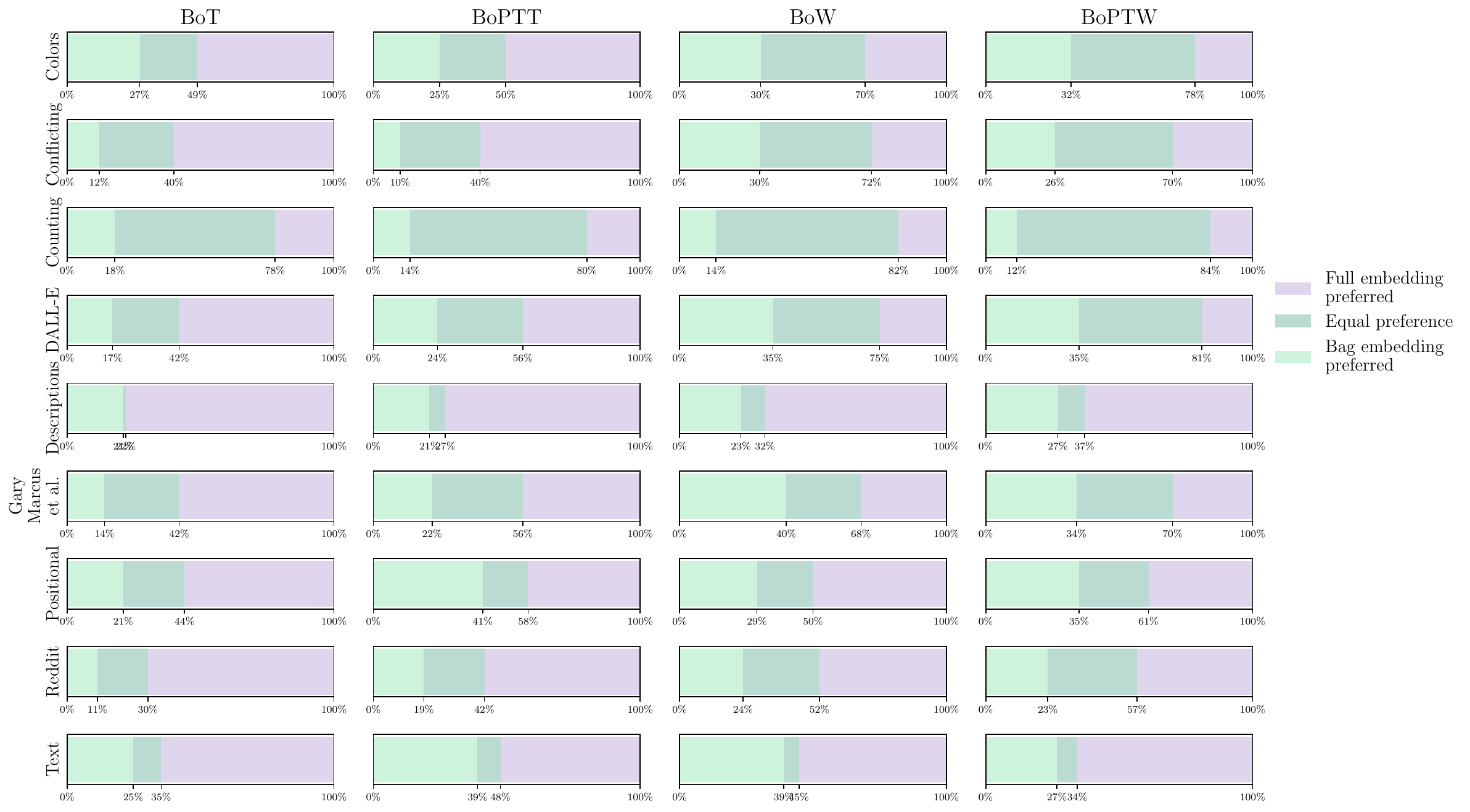}
    \caption{\textbf{{DrawBench categories breakdown for SD3}.}}
    \label{fig:drawbench_sd3}
\end{figure}

\begin{figure}
    \centering
    \includegraphics[width=\linewidth]{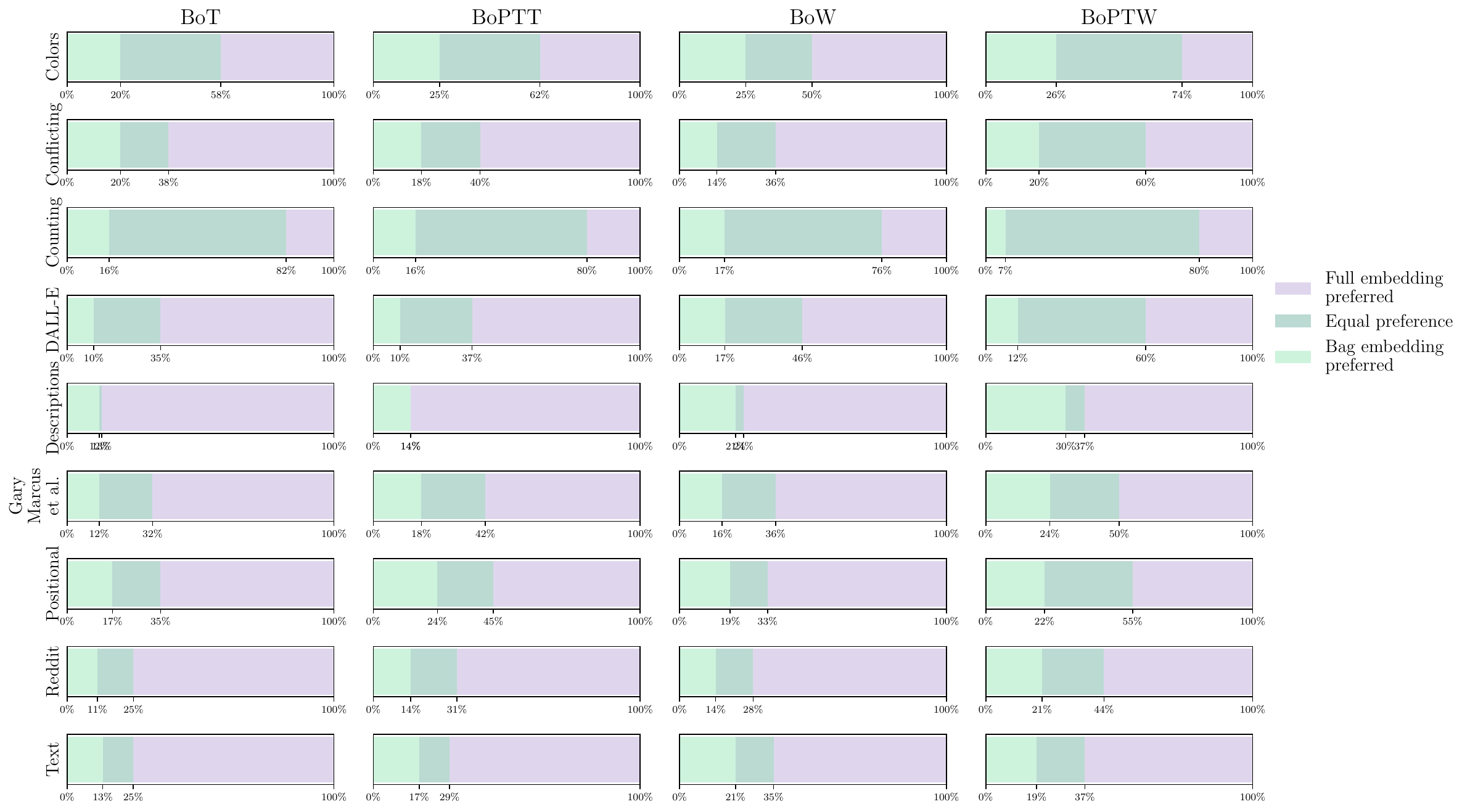}
    \caption{\textbf{{DrawBench categories breakdown for FLUX.1}.}}
    \label{fig:drawbench_flux1}
\end{figure}

\begin{figure}
    \centering
    \includegraphics[width=\linewidth]{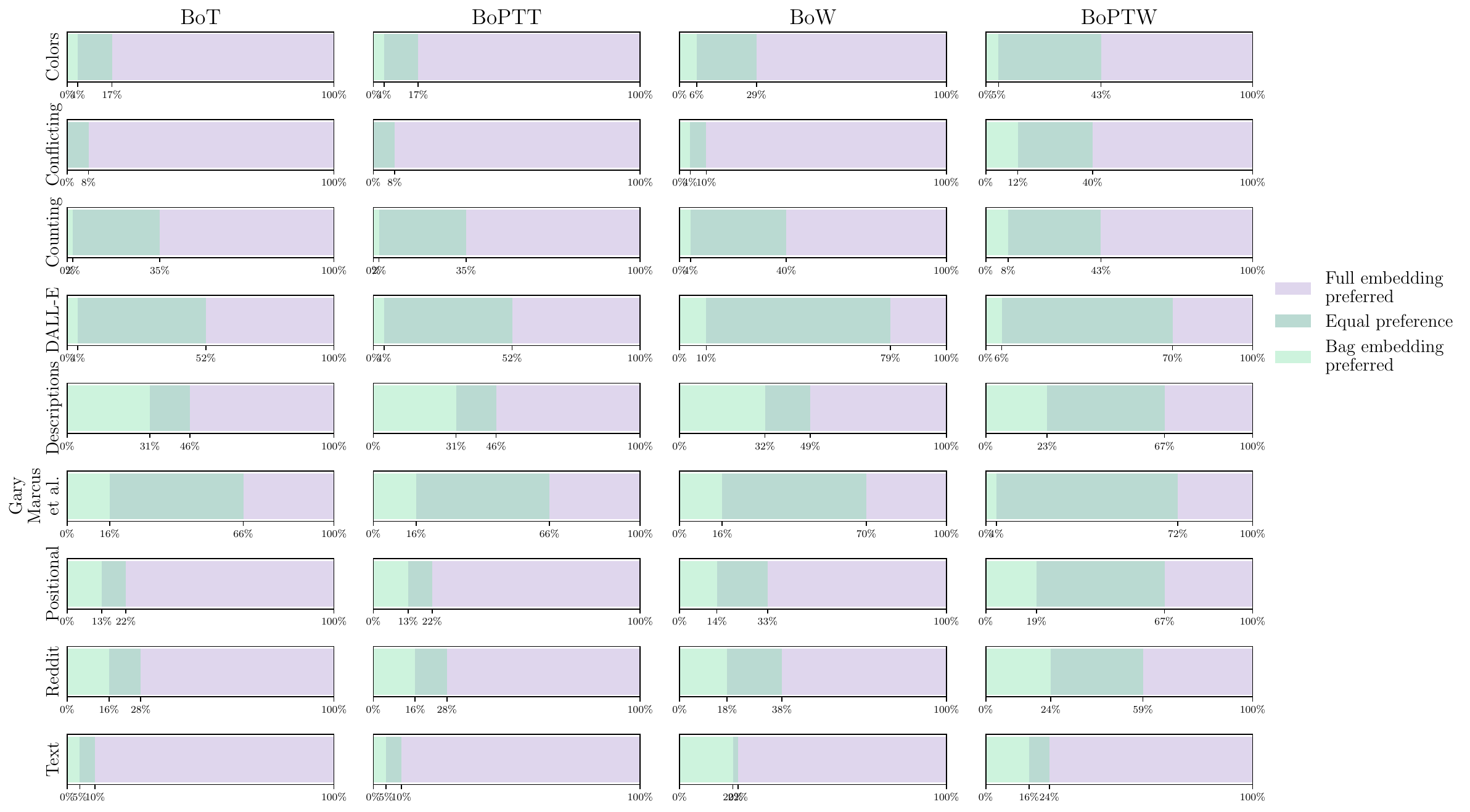}
    \caption{\textbf{{DrawBench categories breakdown for FLUX.2}.}}
    \label{fig:drawbench_flux2}
\end{figure}

\begin{figure}[t]
    \centering
    \includegraphics[width=0.75\linewidth]{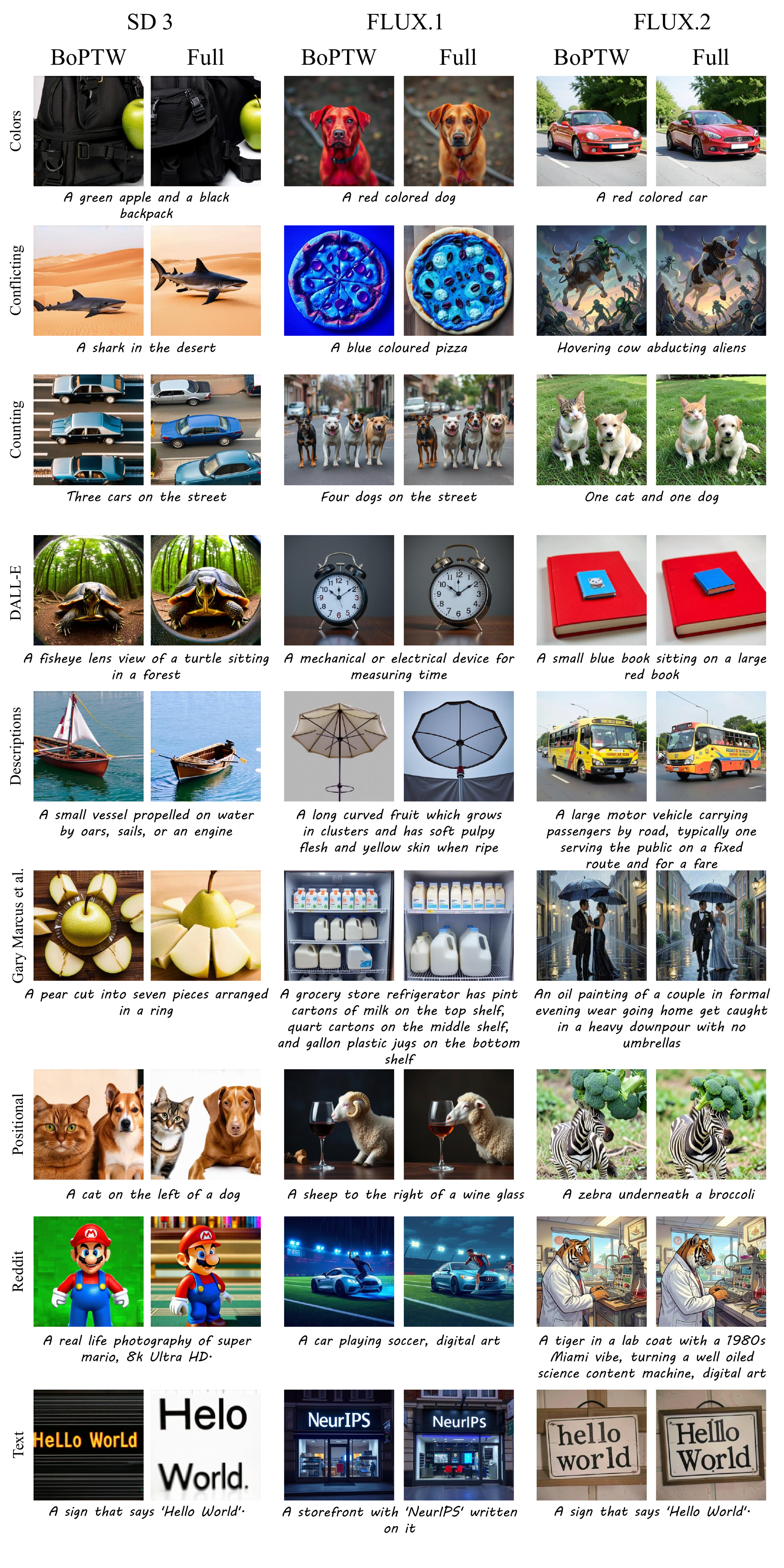}
    \caption{\textbf{Successful prompts from the DrawBench categories.} We present prompts from all DrawBench categories in which the BoPTW embedding was sufficient for the image model to adhere to the prompt.}
    \label{fig:drawbench_supp_good}
\end{figure}

\begin{figure}[t]
    \centering
    \includegraphics[width=0.75\linewidth]{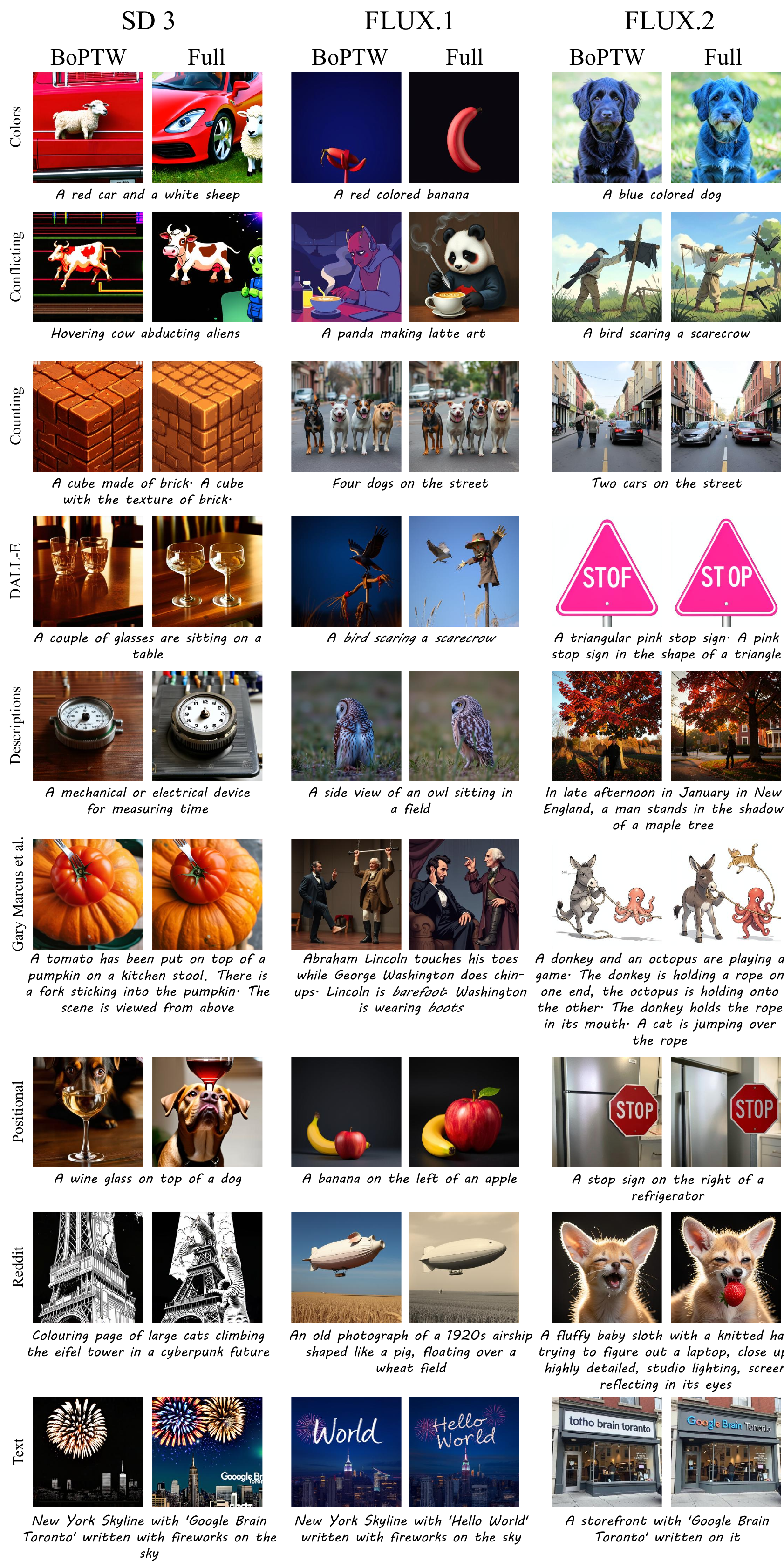}
    \caption{\textbf{Unsuccessful prompts from the DrawBench categories.} We present prompts from all DrawBench categories in which the BoPTW embedding was not sufficient for the image model to adhere to the prompt.}
    \label{fig:drawbench_supp_bad}
\end{figure}

In \cref{fig:drawbench_supp_good}, we provide visual examples across all DrawBench categories for each evaluated model. These examples demonstrate that the BoPTW embeddings are often sufficient for guiding the image model to generate outputs that faithfully adhere to the text.
Conversely, \cref{fig:drawbench_supp_bad} provides visual examples of failure cases where the BoPTW embeddings lack sufficient information, leading the image model to generate images that are semantically misaligned with the input text.
In addition to the category-wise evaluation in~\cref{fig:categories_bars}, \cref{fig:drawbench_sd3,fig:drawbench_flux1,fig:drawbench_flux2} report the text-alignment results for the additional contextless embeddings, showing per-category trends that resemble the overall trend.


\subsection{Evaluation on DPG benchmark}
\label{ssection:Evaluation on DPG benchmark}
The benchmark consists of 1,065 prompts, with an average length of 408 words per prompt. Following the experimental setup described in~\Cref{sec:exp_details}, we generated five images per prompt using both the contextless and full embeddings and evaluated the resulting images with Gemma-3. 
In \Cref{tab:dpg_bncmrk}, we report the non-inferiority rate of contextless embeddings on the DPG benchmark. We observe a consistent trend across all models, with BoPTW outperforming both BoT and BoW. Interestingly, FLUX.1 achieves the best overall performance. 

\begin{table}[ht]
    \caption{Non-inferiority rate in [\%] for DPG benchmark. higher scores        correspond to better results.}
    \label{tab:dpg_bncmrk}
    \small
    \begin{center}
    \begin{tabular}{llll}
    \textbf{Model} & \textbf{FLUX.1} & \textbf{FLUX.2} & \textbf{SD~3} \\
    \textbf{Embedding} &  &  &  \\
    \hline \\
    BoPTW & 57 & 41.5 & 47.2 \\
    BoW & 19.4 & 13.8 & 16.6 \\
    BoT & 10.7 & 7.1 & 9.2 \\
    \end{tabular}
    \end{center}
\end{table}

\subsection{User Study}
\label{ssection:User Study}
We conducted a user study to further compare the text adherence of images generated with the BoPTW embeddings to those generated with the full embeddings. On a random subset of $30$ prompts from DrawBench which are balanced across the TTI models. We asked 25 individuals to rate whether one image adhered significantly better to the textual prompt. We conducted an equivalent study on $30$ prompts from GenEval. On DrawBench, users found no significant advantage on $63\%$ of the prompts, $20\%$ preferred images generated with the full embedding, and $16\%$ preferred images generated with the BoPTW embedding. On the exact same prompts, the judgment of Gemma-3 (as reported in \Cref{sec:exp_details}) was $43\%$, $30\%$, and $27\%$, respectively.On GenEval, the users’ evaluations were $60\%$ (no advantage), $25\%$ (full), and $14\%$ (BoPTW), while the Gemma-3 partition was $47\%$, $40\%$, and $13\%$. When comparing the judgment of Gemma-3 to users’ preferences on these subsets, we find strong alignment: on DrawBench, users agree with the decision of Gemma-3 on $61\%$ of the prompts, which is highly comparable to the rate at which they agree among themselves ($67\%$). Similarly, on GenEval, the user-VLM agreement is $62\%$, compared to a user-user agreement of $73\%$.

\clearpage
\subsection{DiT vs. UNet}
\label{ssection:DiT__UNet}
Following the discussion in the main paper, here we present visual examples of images generated by SD~2.1~\citep{rombach2022ldm} and SDXL~\citep{podell2023sdxl} guided by the BoPTW embeddings. The generated images do not adhere to the prompts at all, and appear to be drawn from a similar distribution, suggesting that the models struggle to infer context and produce prompt-adherent results.

\begin{figure}[ht]
    \centering
    \includegraphics[width=\linewidth]{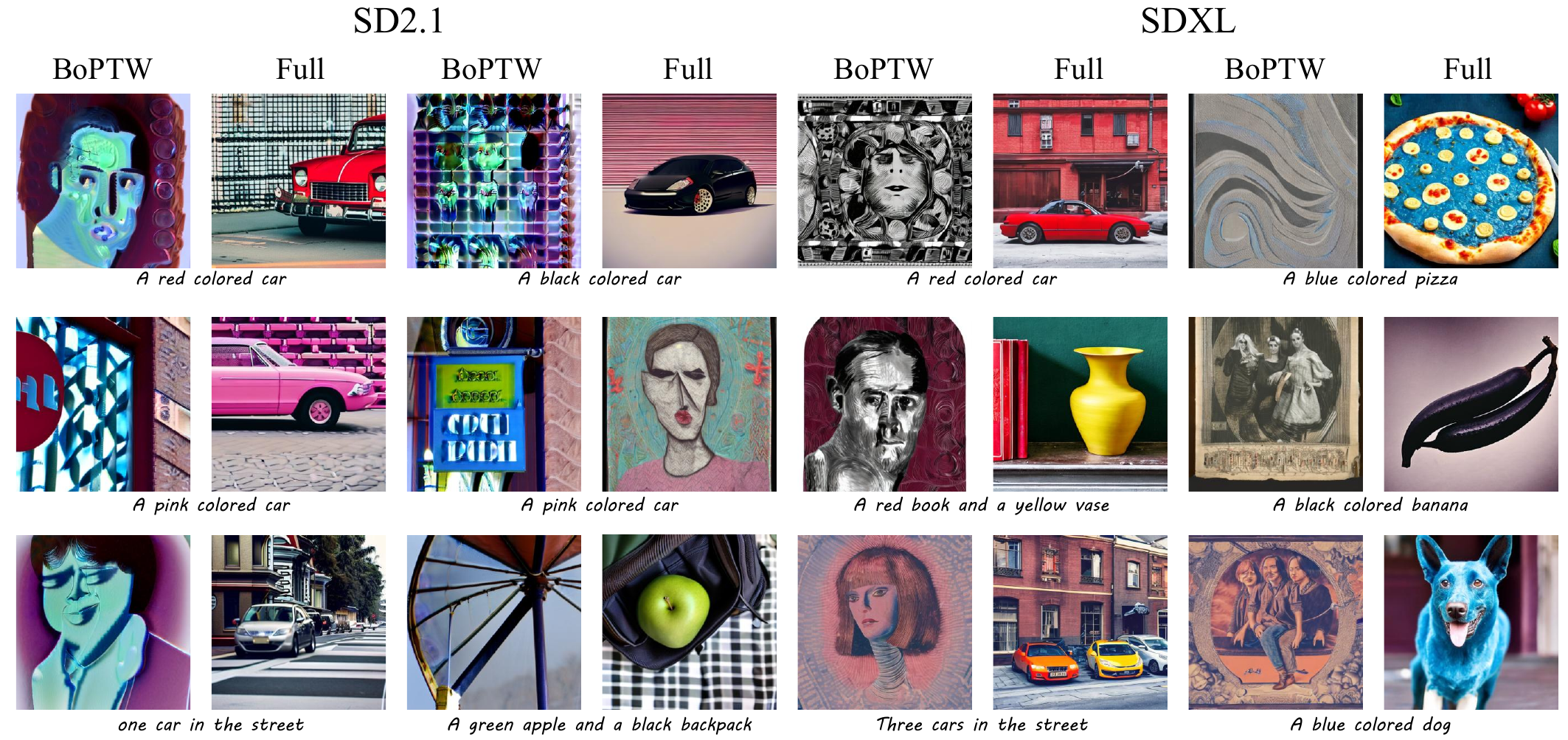}
    \caption{\textbf{Visual examples for contextless-guided generation with SD~2.1 and SDXL}. In each pair, the left image is generated using BoPTW embeddings, while the right is generated using full embeddings. We observe that contextless embeddings are not sufficient to effectively guide image generation.}
    \label{fig:non_dit}
\end{figure}
\clearpage
\section{Experimental details} \label{sec:exp_details}
All experiments were run on an NVIDIA RTX A6000. Image generation with the constructed embeddings is similar to that of the original TTI models.

\subsection{Licenses}\label{sec:license}
Prompt datasets are used to create contextless embeddings and for evaluation: MSCOCO 2014 and 2017 (CC BY 4.0), GenEval (MIT), DrawBench (Apache 2.0) and CC3M (conceptual-captions).
We use five TTI models: SD~2.1 (openrail++), SD~3 (stabilityai-ai-community), SDXL (openrail++), FLUX.1 Schnell (Apache 2.0), and FLUX.2 Klein (Apache 2.0).
We additionally utilize Gemma-3 (Gemma) and Claude-sonnet-4.5 (Proprietary, Anthropic). 

\subsection{Constructing  contextless embeddings}\label{sec:sentences}
As explained in~\cref{sec:exp_setup}, to construct contextless embeddings, we iterate over the training set of CC3M and MSCOCO-2017, collecting all embedding vectors of each token and storing them aside. Then, when iterating over prompts for image generation, for each token that appears less than ten times within these datasets and for each word that is split into multiple sub-tokens, we generate additional sentences using Claude~\citep{anthropic2025claude45}).
We use the following prompt to guide the generation of sentences that should include a given word at a specific token position: 

\begin{boxB}
    You are a prompt generation engine for text-to-image models.
    
    Task:
    Generate 10 high-quality, rational prompts.
    
    Inputs:
    \begin{itemize}
        \item Preserved word: target word
        \item Preserved word position (0-based index): position
        \item Target: text-to-image model
    \end{itemize}

    Rules (STRICT):
    \begin{itemize}
    \item Ensure that the prompts cover diverse and distinct contexts.
    \item Each prompt must be a sequence of space-separated words.
    \item  The preserved word must appear EXACTLY at the given position.
    \item  The preserved word must remain unchanged in spelling and case.
    \item  All other words must be different from the preserved word.
    \item  Do NOT reuse any words from other prompts unless required by grammar.
    \item  Do NOT move the preserved word from its position.
    \item  Prompts must be grammatically valid and semantically coherent.
    \item  Prompts must describe a clear visual scene suitable for image generation.
    \end{itemize}

    Output format (STRICT):
    \begin{itemize}
    \item Output ONLY a valid Python list
    \item The list must contain exactly 10 strings
    \item No explanations, no comments, no extra text
    \end{itemize}
\end{boxB}

\begin{figure}[ht]
    \centering
    \includegraphics[width=\linewidth]{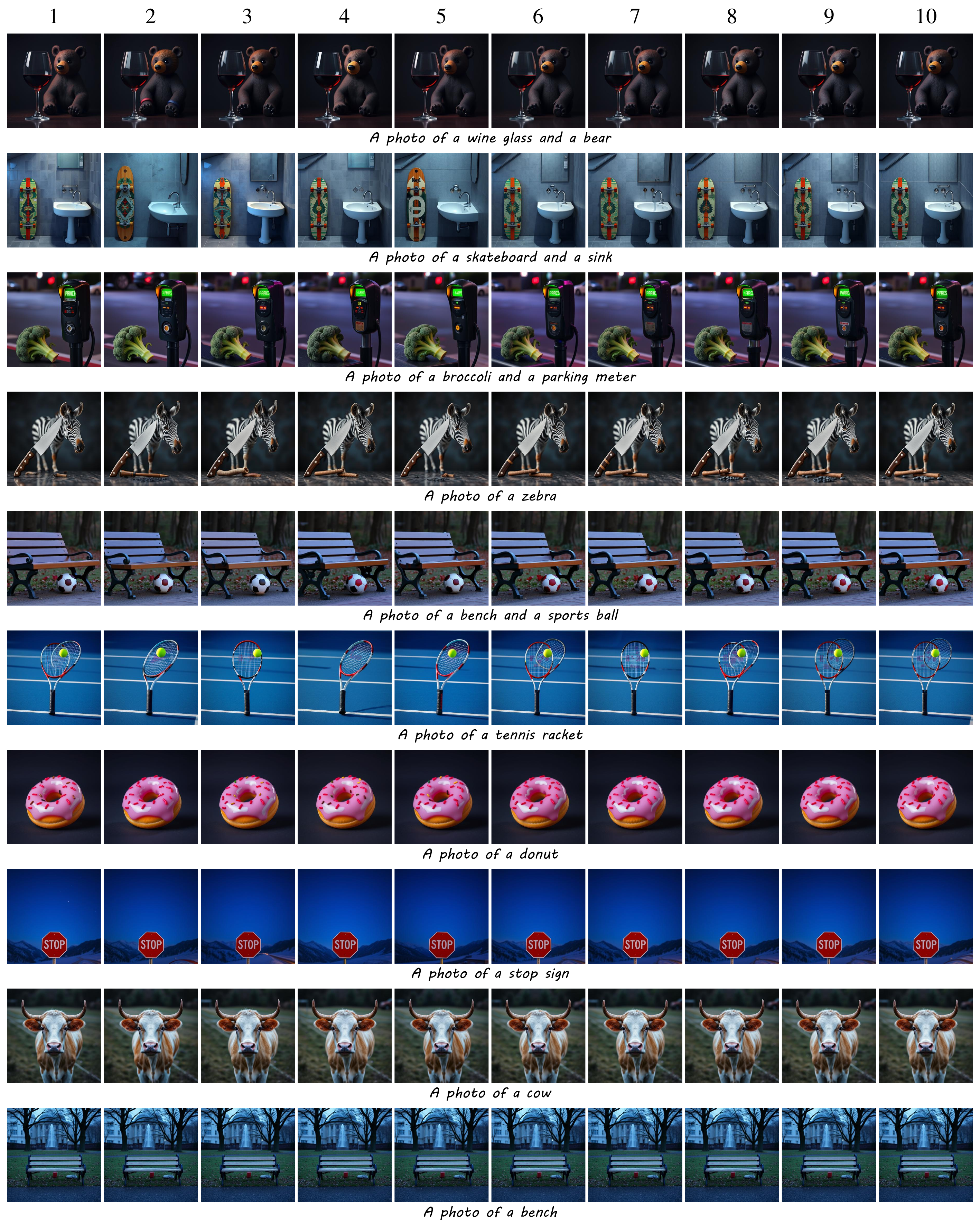}
    \caption{\textbf{Impact of unrelated sentence set size on BoPTW-generations.} We visualize generations for ten GenEval prompts using FLUX.1~Schnell. The number of unrelated sentences used to compute the average embedding is indicated above each column. }
    \label{fig:num_prompts}
\end{figure}

\vspace{1cm}
\clearpage
\myparagraph{Impact of averaging set size.} While using a single unrelated sentence for the contextless embedding of each token in the target prompt is sufficient to provide unrelated context, to ensure a more robust and stable representation.
\Cref{fig:num_prompts} presents BoPTW-guided generations for ten GenEval prompts using FLUX.1 Schnell, with columns representing an increasing number of context-erasing sentences. 
Notably, increasing the size of the averaging pool does not lead to any perceptible semantic change.

\clearpage
\subsection{VLM as a judge evaluation} \label{sec:vlm_eval}
To compare the adherence of the images generated for the full-embedding to those generated for contextless embedding, we employ Gemma-3~\citep{gemmateam2025gemma3technicalreport} as a VLM judge.
To complement the explanation of the three-way comparison described in~\cref{sec:exp_details}, we present the instructions to the VLM.
We implement a two-stage evaluation process where the VLM is given an image pair, one generated with the full embedding and another by one of the contextless embeddings in random order, along with their prompt, and first determines if a significant difference in text adherence exists between an image pair, according to the following prompts.

\vspace{-5pt}
\subsection*{System prompt:}
\vspace{-5pt}
\begin{boxB}
You are an expert image evaluator. Your task is to determine if one image adheres significantly better to a text than another. Answer "Yes" if one is clearly superior, or "No" if they are similar in quality or both are poor. No preambles or postambles.
\end{boxB}

\vspace{-5pt}
\subsection*{User prompt:}
\vspace{-5pt}
\begin{boxB}
Does one image adhere significantly better than the other to the text "$\langle$PROMPT$\rangle$"? Answer with "Yes" or "No".
\end{boxB}

\vspace{0.5cm}
If the model responds "Yes", a second question asks it to identify the superior image. These results are then parsed into the three categories reported in our findings: equal preference for cases where no significant difference was detected, or a specific preference for either the full or contextless embedding.

\vspace{-5pt}
\subsection*{System prompt:}
\vspace{-5pt}
\begin{boxB}
You are an expert image evaluator. Identify which image is superior. Only answer with "first" or "second", no preambles or postambles.
\end{boxB}

\vspace{-5pt}
\subsection*{User prompt:}
\vspace{-5pt}
\begin{boxB}
Which image adheres better to the text "$\langle$PROMPT$\rangle$"? Answer with "first" or "second".
\end{boxB}

\clearpage
\subsubsection{Discussion-cont'} \label{sec:Discussion_app}
To further investigate whether the ability to decode linguistic structure occurs within the image model and specifically in  DiT and not in U-Net, we conduct a controlled experiment using synthetic TTI generation. The goal is to isolate the interaction between text representation and the image model. We construct a synthetic dataset of $10,000$ $3 \times 32 \times 32$ images featuring solid-colored geometric shapes (circle, square, triangle) rendered on a uniform background without spatial occlusion. Each image is paired with a caption specifying the attributes (red, blue, green, yellow) and spatial relations (left of, right of) for two objects, such as ``a red circle left of a blue square''. This yields $288$ unique prompts, requiring the model to jointly represent object identity, attribute binding, and simple compositional syntax.
We train two conditional generative models of comparable capacity ($~30M$ parameters): a convolutional U-Net (utilizing text cross-attention across all resolution levels) and a diffusion Transformer (DiT). 
Both models receive representations from a frozen T5-XXL text encoder. 
Models are trained using a Rectified Flow matching objective with continuous timesteps $t \sim U(0,1)$, a batch size of $128$, and the AdamW optimizer for $10,000$ iterations.
At inference, the ODE is integrated using $50$ discrete Euler steps. To evaluate text adherence using a VLM judge, the $32 \times 32$ generated images are upsampled to $256 \times 256$ via nearest-neighbor interpolation. We then compare images generated with the full embeddings against those generated with the BoPTW embeddings using Gemma-3 as the judge, following the protocol described in  \Cref{sec:exp_details}. 
The DiT architecture demonstrated a non-inferiority rate of $67.56\%$, substantially outperforming the U-Net, which achieved just $26.27\%$.

\begin{figure}[h]
    \centering
    \includegraphics[width=\linewidth]{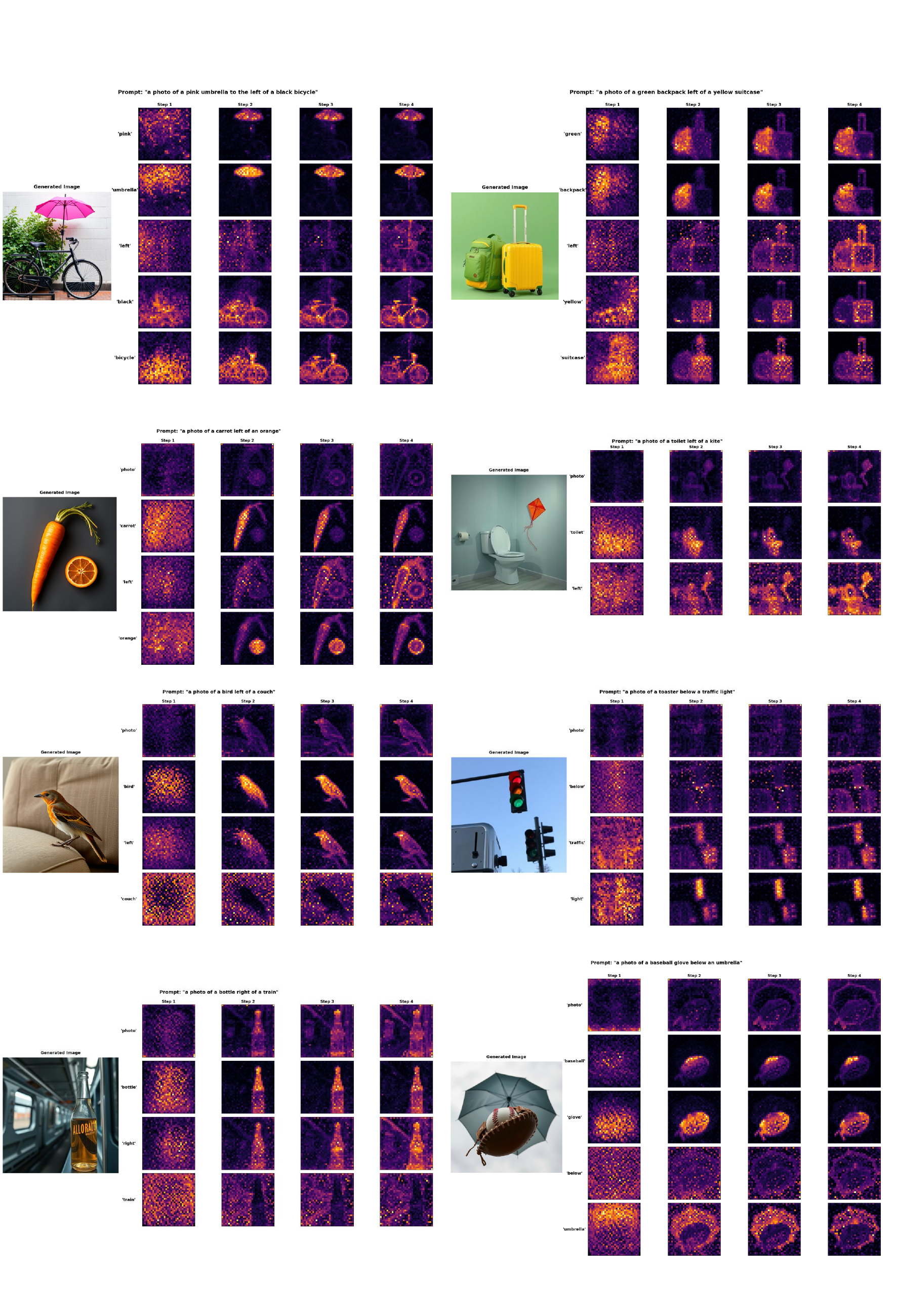}
    \caption{\textbf{Cross-Attention maps.} By extracting and visualizing spatio-temporal cross-attention trajectories from the double-stream blocks in FLUX.1, we demonstrate that linguistic decoding is performed internally by the image model. Each panel displays the generated image on the left, accompanied on the right by the token-specific attention maps obtained across the denoising steps.}
    \label{fig:mech}
\end{figure}

\paragraph{Mechanistic evidence of internal linguistic decoding.}
To substantiate our behavioral findings with internal mechanistic evidence, we investigate the spatio-temporal evolution of attention within the DiT architecture during the generation process. We extract the explicit joint attention matrices from the double-stream transformer blocks of the FLUX.1 model. By isolating the attention map of specific tokens (\eg the token ``green'' in the prompt ``a photo of a green backpack left to a yellow suitcase'') and projecting it back onto the spatial latent grid, we can track how the model resolves linguistic structure across the ODE integration timesteps. As illustrated in \Cref{fig:mech}, the cross-attention trajectories under the BoPTW embeddings condition reveal the model's internal computational dynamics. In the early stages of generation, the attention maps are highly diffuse, primarily reflecting the unstructured noise inherent to the initial latents. However, as visual features begin to emerge, the DiT's internal joint-attention mechanisms actively process the absolute positional tags to resolve semantic ambiguities. Despite the BoPTW embeddings lacking any explicit relational links between the attributes and the objects, the attention mass converges onto the correct spatial coordinates. This dynamic spatial localization provides direct mechanistic evidence that the image model independently solves complex linguistic structures within its own.

\clearpage
\subsubsection{VLM agreements} \label{sec:vlm_agreement}
In \Cref{tab:agreement}, we report the agreement between the two VLMs across all models and datasets. 
The overall agreement rate across $300K$ pairs of images (all TTI models, benchmarks, contextless embeddings, and multiple seeds per prompts) is $70\%$.

\begin{table}[h]
\centering
\small
\setlength{\tabcolsep}{3pt}
\begin{tabular}{lllrrrrrrrrr}
TTI Model & Dataset & VLM &
\multicolumn{3}{c}{BoT} &
\multicolumn{3}{c}{BoW} &
\multicolumn{3}{c}{BoPTW} \\
&&&
Preferred & No pref. & Full &
Preferred & No pref. & Full &
Preferred & No pref. & Full \\
FLUX.1 & DrawBench & Gemma & 14.6 & 24.6 & 60.8 & 17.9 & 23.5 & 58.6 & 21.0 & 34.9 & 44.1 \\
FLUX.1 & DrawBench & Qwen  & 7.4  & 32.1 & 60.5 & 8.3  & 29.1 & 62.6 & 14.0 & 44.9 & 41.2 \\
FLUX.1 & GenEval   & Gemma & 15.9 & 37.0 & 47.1 & 19.3 & 39.7 & 41.0 & 18.4 & 55.3 & 26.4 \\
FLUX.1 & GenEval   & Qwen  & 8.0  & 46.2 & 45.8 & 11.4 & 49.5 & 39.1 & 14.9 & 67.3 & 17.8 \\
FLUX.1 & MSCOCO    & Gemma & 17.8 & 37.3 & 44.9 & 18.1 & 38.8 & 43.0 & 17.8 & 47.1 & 35.1 \\
FLUX.1 & MSCOCO    & Qwen  & 14.2 & 40.1 & 45.7 & 14.7 & 41.7 & 43.6 & 15.9 & 50.3 & 33.8 \\

FLUX.2 & DrawBench & Gemma & 11.2 & 19.8 & 69.0 & 14.8 & 26.0 & 59.2 & 14.8 & 39.2 & 46.0 \\
FLUX.2 & DrawBench & Qwen  & 2.1  & 14.0 & 83.9 & 9.0  & 31.9 & 59.2 & 10.2 & 42.0 & 47.8 \\
FLUX.2 & GenEval   & Gemma & 3.4  & 30.0 & 66.5 & 3.6  & 40.4 & 56.0 & 3.1  & 63.8 & 33.1 \\
FLUX.2 & GenEval   & Qwen  & 1.3  & 32.6 & 66.1 & 1.4  & 41.9 & 56.7 & 2.5  & 65.3 & 32.2 \\
FLUX.2 & MSCOCO    & Gemma & 11.7 & 31.3 & 57.0 & 11.6 & 45.2 & 43.2 & 8.9  & 57.6 & 33.5 \\
FLUX.2 & MSCOCO    & Qwen  & 6.4  & 29.8 & 63.7 & 7.5  & 43.0 & 49.4 & 6.9  & 55.7 & 37.5 \\

SD3 & DrawBench & Gemma & 18.6 & 23.7 & 57.7 & 28.3 & 30.9 & 40.8 & 27.6 & 35.1 & 37.2 \\
SD3 & DrawBench & Qwen  & 8.4  & 30.7 & 61.0 & 18.7 & 39.6 & 41.7 & 19.0 & 43.6 & 37.4 \\
SD3 & GenEval   & Gemma & 22.6 & 21.3 & 56.1 & 27.9 & 40.8 & 31.3 & 26.5 & 42.1 & 31.4 \\
SD3 & GenEval   & Qwen  & 13.5 & 31.6 & 54.9 & 19.9 & 54.3 & 25.9 & 20.7 & 56.1 & 23.2 \\
SD3 & MSCOCO    & Gemma & 20.5 & 25.9 & 53.6 & 21.8 & 28.2 & 50.1 & 22.3 & 48.2 & 29.5 \\
SD3 & MSCOCO    & Qwen  & 15.2 & 28.3 & 56.5 & 18.3 & 30.9 & 50.8 & 19.9 & 50.2 & 29.9 \\

\end{tabular}

\caption{\textbf{Agreement between Gemma-3 and Qwen3.} 
We report the results obtained by each VLM across all TTI models and datasets. ``Preferred'' indicates a preference for the contextless embedding, ``No pref.'' indicates no significant preference, and ``Full'' indicates a preference for the full embedding. All values are reported as percentages of prompts.}
\label{tab:agreement}
\end{table}
\FloatBarrier

\end{document}